\documentclass{article}

\usepackage[eandd, final]{neurips_2026}

\usepackage[utf8]{inputenc} 
\usepackage[T1]{fontenc}    
\usepackage{hyperref}       
\usepackage{url}            
\usepackage{booktabs}       
\usepackage{amsfonts}       
\usepackage{nicefrac}       
\usepackage{microtype}      
\usepackage{xcolor}         
\usepackage{amsmath}
\usepackage{amssymb}
\usepackage{graphicx}
\usepackage{tikz}
\usetikzlibrary{positioning}

\usepackage{listings}
\usepackage{textcomp} 
\usepackage{comment}

\lstdefinestyle{promptstyle}{
  basicstyle=\ttfamily\footnotesize,
  breaklines=true,
  breakatwhitespace=true,
  frame=single,
  columns=fullflexible,
  keepspaces=true,
  showstringspaces=false,
  upquote=true,
  backgroundcolor=\color{gray!4},
  xleftmargin=0.5em,
  xrightmargin=0.5em,
  aboveskip=0.75em,
  belowskip=0.75em
}

\title{The Labeling Problem in Hallucination Detection Benchmarks: An Empirical Evaluation}

\author{%
  Jorma Valjakka
  \\
  Department of Computer Science\\
  University of Helsinki\\
  \texttt{jorma.valjakka@helsinki.fi} \\
  \And
  Juhani Kivimäki \\
  Department of Computer Science\\
  University of Helsinki\\
  \texttt{juhani.kivimaki@helsinki.fi} \\
  \And
  Juha Mylläri \\
  Department of Computer Science\\
  University of Helsinki\\
  \texttt{juha.myllari@helsinki.fi} \\
  \And
  Jukka K. Nurminen \\
  Department of Computer Science\\
  University of Helsinki\\
  \texttt{jukka.k.nurminen@helsinki.fi} \\
}

\begin{document}

\maketitle

\begin{abstract}
In recent years, several methods for detecting when large language models (LLMs) hallucinate have been developed. These methods are often benchmarked with open-domain question answering (QA) datasets containing questions and corresponding short reference answers. First, an LLM is used to generate answers to questions within the QA dataset. Then, some automated labeling strategy is used to label these answers as hallucinated or not by comparing them with the reference answers in the dataset. This evaluation setting creates a methodological ambiguity between two criteria: reference faithfulness --- whether the answer is fully supported by the reference --- and factual correctness --- whether the answer is free from contradictions and factually false specific claims. In practice, automated labelers may apply the former criterion even when the intended target is the latter.

We study this potential criterion mismatch using 900 human-labeled question--answer pairs spanning three commonly used QA datasets and three generator models, with labels targeting answer-level factual correctness. As automated label generation strategies, we evaluate lexical similarity metrics, a reference-entailment NLI baseline, and seven LLM judges under controlled prompt variants. Our experiments reveal substantial disagreement both among different automated label-generation strategies and between these automated labels and human annotations. Many strategies also exhibit strong directional error biases, and for most judge--generator pairs, replacing a faithfulness-oriented prompt with a factual-correctness prompt improves agreement with human annotations and reduces false-positive dominance, indicating that automated hallucination labels depend strongly on how the target criterion is specified. Label-source choice should therefore be considered a fundamental part of benchmark design and made explicit, validated, and matched with the benchmark goal.
\end{abstract}

\section{Introduction}
\label{sec:introduction}

Large language models (LLMs) are notoriously prone to hallucinations. That is, they generate fluent statements containing fabricated or factually incorrect information~\citep{ji2023survey,niu2024ragtruth}. Currently, new methods for hallucination detection are constantly being developed. Many of these methods are benchmarked using common question-and-answer (QA) datasets that contain questions and a set of reference answers to each question~\citep{Yin_Lid, NEURIPS2024_Semantic_Density,chen2024inside, duan-etal-2024-shifting, farquhar2024detecting}. Typically, the answer-label pairs needed for benchmarking these methods are produced by using an LLM to generate answers to the questions in the QA dataset and applying some automated strategy to label them as hallucinated or not~\citep{bang-etal-2025-hallulens,janiak2025illusion}. However, the validity of this approach hinges on a critical, often overlooked question: do these labels actually measure the intended construct? Common automated label sources, including lexical similarity metrics~\citep{lin2004rouge,zhang2019bertscore} and LLM judges, are often used without directly validating this construct alignment~\citep{janiak2025illusion,kulkarni2025mirage}, which might have a considerable effect on the evaluation outcomes.

We study the conceptual discrepancy between reference faithfulness and factual correctness in open-domain QA. A reference-faithfulness labeler treats answer content not entailed by the reference as hallucinated. A factual-correctness labeler treats an answer as hallucinated only if it contains a contradiction or an untrue specific claim. The two criteria diverge when an answer is correct but adds true information that is absent from the reference answer set. In such cases, strict faithfulness can systematically penalize correct elaborations. Prior to our work, this criterion mismatch has not been systematically studied as a source of automatic-label error in open-domain QA hallucination benchmarks to the best of our knowledge. Our work is not intended as a general critique of faithfulness: in evidence-grounded tasks such as RAG and summarization~\citep{niu2024ragtruth,maynez2020faithfulness}, faithfulness is often the intended target. Our claim focuses strictly on open-domain QA benchmarks that target factual correctness, where the labeling criterion must be explicit and validated against that specific target.

To ground our analysis, we evaluate lexical similarity metrics, a reference-entailment NLI baseline, and seven diverse LLM judges (spanning OpenAI, Anthropic, and local open-weight models) against newly collected human annotations. We analyze the inter-annotator agreement among different automated labeling strategies and, in particular, with a human annotator. We find that the overall agreement between all automated methods and a human annotator is substantially lower than that between two human annotators. Furthermore, we examine the false-positive and false-negative error structures across label sources when human-annotator labels are treated as the ground truth. We also compare these label sources under controlled prompt ablations to isolate the impact of different labeling criteria.

In this paper, we make the following contributions.
\begin{itemize}\itemsep0pt
    \item We construct a 900-pair human-annotated evaluation sample for auditing automated hallucination labelers in open-domain QA, with strong independent inter-annotator agreement on a 300-item overlap subset.
    \item We formalize the distinction between reference faithfulness and factual correctness, and show empirically that label sources approximating strict faithfulness produce false-positive-dominant errors against a factual-correctness benchmark.
    \item We conduct a controlled prompt-ablation study showing that
    replacing a faithfulness-style criterion with a factual-correctness criterion substantially reduces this bias across most judges. Frontier judges from different model families show that capability alone does not determine the size of faithfulness-induced bias.
    \item We analyze different label sources in false-positive/false-negative error space, showing that lexical metrics, strict reference-entailment NLI, faithfulness-style prompts, and criterion-aligned prompts occupy distinct parts of the error space.
\end{itemize}

\section{Related Work}
\label{sec:related}

\looseness=-1
``Hallucination'' does not mean exactly the same thing across benchmarks~\citep{bang-etal-2025-hallulens}. Some datasets emphasize factual correctness with respect to world knowledge~\citep{li2023halueval}, while others focus on faithfulness to a provided source or retrieved evidence, as in summarization and RAG settings~\citep{maynez2020faithfulness,niu2024ragtruth}. Related benchmark frameworks such as \textsc{TRUE}~\citep{honovich2022true} standardize multiple factual-consistency datasets from diverse tasks under a common evaluation protocol, enabling example-level comparison of factual-consistency metrics across task settings. Factuality evaluation is sensitive to task formulation, dataset design, and the specific criterion being operationalized~\citep{pagnoni2021understanding, kulkarni2025mirage}. Therefore, findings from different hallucination benchmarks or labeling protocols may not be directly comparable without accounting for differences in the underlying construct being measured.

The proposed detectors evaluated in the benchmark studies cited in Section~\ref{sec:introduction} do not use benchmark reference answers as inputs. Instead, they rely on uncertainty or semantic relationships among model-generated samples \citep{farquhar2024detecting,duan-etal-2024-shifting}, semantic density estimated from model-generated reference responses \citep{NEURIPS2024_Semantic_Density}, or internal representations \citep{chen2024inside, Yin_Lid}. Benchmark gold answers are used separately to construct the correctness labels against which the detector scores are evaluated.

The QA label rules are not uniform. \citet{duan-etal-2024-shifting} either use ROUGE-L or sentence-embedding similarity above 0.5; \citet{NEURIPS2024_Semantic_Density} use ROUGE-L greater than 0.3; \citet{chen2024inside} report both ROUGE-L greater than 0.5 and cosine similarity greater than 0.9 as separate correctness measures; and \citet{Yin_Lid} use ROUGE-L at least 0.5 in their main experiments, with ROUGE-L at least 0.3 and NLI entailment as sensitivity analyses. These rules compare answers with benchmark references and do not explicitly require evidence beyond them. 
\citet{farquhar2024detecting} use a different label source for sentence-length QA: GPT-4 is prompted to decide whether the proposed answer means the same as the expected answer, without separately verifying its
factual correctness.
As such, we consider all of the above to operationalize reference-faithfulness in their labeling strategies. However, the 
paragraph-length \textsc{FactualBio} evaluation  in~\citet{farquhar2024detecting} differs: GPT-4 decomposes the generated biographies into individual factual claims, which the authors then manually label as true or false.

A related but distinct line of work moves from whole-output scoring to more explicit verification. Claim-level and knowledge-centric pipelines decompose model outputs into atomic claims and check them individually,
enabling finer-grained factual assessment than single score estimates over entire responses~\citep{min2023factscore,hu2024knowledgecentric}. These approaches clarify the unit of evaluation but not the criterion used
to define hallucination in a benchmark. In particular, making evaluation more fine-grained does not eliminate possible mismatch
between reference faithfulness and factual correctness when hallucination labels are defined for a specific benchmark.

Recent meta-evaluation work shows that automatic evaluation signals can diverge sharply from human judgments at the example level. In summarization and related factual consistency settings, lexical overlap metrics such as ROUGE-L and semantic similarity metrics such as BERTScore have shown limited validity as measures of faithfulness or factuality relative to human
annotations~\citep{maynez2020faithfulness,honovich2022true,pagnoni2021understanding}. QA-specific meta-evaluations support this: automatic metrics can overstate progress or misrank methods when the evaluation signal is
poorly aligned with human factual
judgments~\citep{janiak2025illusion,kulkarni2025mirage}. We therefore focus on label-source validity, that is, whether the labels used to
evaluate hallucination detectors actually reflect the intended construct.

LLM-as-a-judge methods offer a scalable alternative to fixed lexical similarity metrics, but prior work shows that their reliability depends strongly on evaluation design. Judge outputs can vary with model
capability, prompt formulation, and related design
choices~\citep{zheng2023mtbench,thakur2025judgingjudges}. Our work builds on this literature, but asks a more specific question for
open-domain QA with short reference answers: whether disagreement arises not only because judges are noisy or weak, but because the
evaluation criterion encoded in a prompt is misaligned with the target construct. We study this as construct-level misalignment between a
reference-faithfulness label source and a benchmark target of factual correctness, and show that it yields a strongly directional false-positive error pattern in automated
labeling.

\section{Methods}
\label{sec:method}

In this section, we define the hallucination constructs we study and formalize two competing labeling criteria: faithfulness to a provided reference versus factual correctness under human adjudication. We then describe our dataset generation and human-annotated benchmark, followed by the automated label sources and evaluation protocol used to compare judges, quantify agreement with humans, and analyze error direction and bias.

\subsection{Labeling Constructs: Faithfulness vs.\ Factual Correctness}
\label{sec:constructs}

Let $q$ denote a question, $a$ an answer produced by an LLM, and $R=\{r_1,\ldots,r_k\}$ a set of reference answers given in a dataset. We use $\mathrm{claims}(a)$ as an abstract decomposition of the factual content of $a$ into individual claims $c$; this is a construct-level idealization, not a literal preprocessing step. We write $R \models c$
if the reference provides sufficient
evidence for claim $c$, and $R \models \neg c$ if it contradicts $c$. Note that $R \not\models c$ is strictly weaker than
$R \models \neg c$: a claim can be unsupported by the reference without being contradicted by it. This distinction is central to the conceptual discrepancy we study.

Let $\mathrm{verifiable}(c)$ indicate that $c$ is a specific factual claim (e.g., a date, name, number, or entity assertion), and
$\mathrm{true}(c)$ that $c$ is judged factually correct under our annotation protocol; the predicate denotes correctness under that protocol, not an assumed external oracle (see Appendix~\ref{app:annotation}). We formalize the two labeling criteria as indicator functions, where 1 signifies a hallucination:
\begin{align}
y_{\text{faith}}(a,R)
  &= \mathbf{1}\!\left[\exists c \in \mathrm{claims}(a)\;:\;
     R \not\models c\right] \\
y_{\text{fact}}(a,R)
  &= \mathbf{1}\!\left[\exists c \in \mathrm{claims}(a)\;:\;
     R \models \neg c \;\vee\;
     (\mathrm{verifiable}(c) \wedge \neg\mathrm{true}(c))\right]
\end{align}
The faithfulness criterion triggers if any claim is unsupported by the reference; the factual-correctness criterion triggers only on
contradictions or untrue factual claims. In practice, contradiction is instantiated as direct conflict with reference-asserted facts, not as formal logical negation. Thus, the criteria agree on three canonical cases --- a correct core answer alone (both 0), a correct core with a false added specific detail (both 1), and a wrong core answer that contradicts the reference (both 1) --- but diverge on a fourth: a correct core answer with a true elaboration absent from the short reference, where $y_{\text{faith}}{=}1$ but $y_{\text{fact}}{=}0$. Relative to a factual-correctness criterion, a labeler that operationalizes $y_{\text{faith}}$ might therefore label factually correct elaborations absent from the reference as hallucinations, producing false positives for this class of cases. In the standard source-relative taxonomy~\citep{maynez2020faithfulness,ji2023survey}, $R \models \neg c$ is a canonical intrinsic-hallucination case, whereas $R \not\models c$ together with $R \not\models \neg c$ is a source-extrinsic addition; the predicate $\mathrm{true}(c)$ introduces a separate world-truth axis over that class, so that ``extrinsic to the source'' and ``factually incorrect'' are not synonymous. Note that $y_{\text{fact}}$ is answer-level rather than core-answer-level: a correct core answer does not override an incorrect specific detail elsewhere in the response.

\subsection{Dataset Generation}

We drew candidate questions from the validation splits of TriviaQA~\citep{joshi2017triviaqa}, HotpotQA~\citep{yang2018hotpotqa}, and TruthfulQA~\citep{lin2021truthfulqa}. These datasets were selected to represent three different QA settings: factoid questions in TriviaQA, multi-hop questions in HotpotQA, and questions designed to elicit answers based on common misconceptions in TruthfulQA. 

For each selected question, we generated one answer from each of three open-weight models --- Llama-3-8B~\citep{llama3modelcard},
Gemma-2-9B~\citep{gemma_2024}, and
Mistral-7B~\citep{mistral2024instructv03} --- under shared sampling settings (see Appendix~\ref{app:generation}). We chose these open-weight models to represent diverse model families while keeping answer generation tractable. Sampling was performed at the question level rather than at the question--answer-pair level: once a question was selected, all three generator outputs for that question were included. Using the same question IDs across generators supports paired cross-generator comparisons.

\subsection{Human-Annotated Benchmark}
\label{sec:benchmark}

Our human-labeled hallucination annotation protocol targets $y_{\text{fact}}$. 
Annotators were instructed to follow a reference-first procedure. This meant that they initially compared the answer against the reference answer set. If this comparison was insufficient to adjudicate a specific factual claim, annotators were allowed to consult external sources (see Appendix~\ref{app:annotation} for more details).



For the human-annotated benchmark, we randomly sampled 300 questions, allocated approximately equally across the three datasets (TriviaQA: 100, HotpotQA: 101, TruthfulQA: 99), each paired with answers from all three generator models, yielding $300 \times 3 = 900$ question--answer pairs. Sampling at the question level and retaining all three generator answers supports paired comparisons across generators while capturing variation in how different models answer the same question. The three datasets further provide variation across short factoid, multi-hop, and misconception-eliciting QA settings. We treat this benchmark as a diagnostic audit set for comparing automated label sources on fixed human-labeled items, not as a population-level estimate of hallucination prevalence.

All 900 question--answer pairs were labeled by a primary annotator (A1) under the protocol described in Appendix~\ref{app:annotation}. To estimate reliability, a second annotator (A2) independently labeled a subset of 300 question--answer pairs, where the questions were stratified with approximately equal allocation across the three datasets, and for each question, one answer from each of the three generators was included. The resulting agreement is summarized in Table~\ref{tab:iaa} in Appendix~\ref{app:annotation}. Using Cohen's $\kappa$~\citep{cohen1960coefficient} as the inter-annotator agreement statistic,
we found the independent inter-annotator agreement to be strong ($\kappa=0.808$, 95\% CI: [0.729, 0.879]; see Appendix~\ref{app:annotation}).

All downstream evaluations used the complete A1 annotation set, \texttt{annotator1\_label}\footnote{We share the dataset and RAI documentation via Harvard Dataverse: \url{https://doi.org/10.7910/DVN/PCHISZ}.}, as the human reference labels for the 900 question--answer pairs. A2 labels were used only to estimate annotation reliability and to support quality control. Hallucination prevalence under the A1 protocol was 44.7\% for Llama-3-8B, 38.0\% for Gemma-2-9B, and 51.0\% for Mistral-7B. At the dataset level, the hallucination rate was 72.3\% for HotpotQA, 30.3\% for TriviaQA, and 30.6\% for TruthfulQA.

\subsection{Automated Label Sources}
\label{sec:label_sources}

We evaluated three families of automated labelers.

\paragraph{Lexical similarity metrics.}
We used ROUGE-L~\citep{lin2004rouge} and
BERTScore~\citep{zhang2019bertscore} to represent lexical overlap and embedding-based reference similarity, respectively. Additional metrics are included in Appendix~\ref{app:new_metrics}. For each generator, threshold $\tau$ was swept from 0 to 1 in steps of 0.01, with direction fixed so that lower similarity scores indicated hallucination. We report two thresholding protocols: an oracle-$\tau$ and a 3-fold cross-validated $\tau$. The oracle-$\tau$ maximizes agreement with human-annotated labels on the full evaluation set. In the cross-validated protocol, $\tau$ is selected on the two training folds and applied to the held-out fold; the reported labels concatenate the three out-of-fold predictions. The oracle-$\tau$ serves as an optimistic diagnostic upper bound rather than a deployable procedure. The CV protocol is used as the primary non-oracle estimate.

\paragraph{Reference-entailment NLI baseline.}
We evaluated a T5-11B model trained similarly to the NLI baseline TRUE~\citep{honovich2022true}, fine-tuned on a mixture of NLI datasets (see Appendix~\ref{app:nli-fine-tuning-datasets}). We use this model with the reference as premise and the model answer as hypothesis, labeling a prediction as a hallucination if the entailment probability falls below $\tau{=}0.5$. We refer to this strategy as the \textit{strict} setting. For comparability with these similarity metrics, we also report an oracle best-$\tau$ variant. Both settings operationalize $y_{\text{faith}}$: entailment is defined relative to the premise, so neither can distinguish a claim that is unsupported by the reference but factually true from one that is false.

\paragraph{LLM judges.}
We evaluate seven LLM judges spanning proprietary API models and local open-weight models. The primary ablation judge is
GPT-5-mini~\citep{openai_gpt5mini}, evaluated under three prompt variants: p1 (faithfulness-style criterion), p2 (factual-correctness), and p3 (extended version of p2). Full prompt texts for these variants are given in Appendix~\ref{app:judge_prompts}.

To disentangle the effect of labeling criterion from prompt structure, we test two additional prompts with GPT-5-mini alongside p1 and p2: p1s, a structured faithfulness prompt, and p2t, a terse factual-correctness prompt. The auxiliary prompts and full details of this experiment are provided in Appendix~\ref{app:factorial_ablation}.

To test transfer across judge capability and model family, we also evaluate GPT-5-nano~\citep{openai_gpt5nano}, the OpenAI frontier model GPT-5.4~\citep{openai_gpt54}, and the Anthropic frontier model Claude Opus 4.7~\citep{anthropic_opus47}. The two frontier model judges probe whether criterion alignment is determined by capability alone or also varies across model families. Judge identifiers, providers, and
experimental roles are summarized in Appendix~\ref{app:judge_setup},
Table~\ref{tab:judge_setup}.

We also evaluate three open-weight local judges: Gemma-2-9B,
Llama-3-8B, and Qwen2.5-7B~\citep{qwen2.5}, with the same p1/p2/p3 prompts on all
three generator sets. Qwen2.5-7B is not used as a generator, so it
serves as an independent judge. Self-evaluation cases arise when
judge and generator coincide (Gemma judging Gemma, Llama judging
Llama); we mark these explicitly in the results and do not rely on
them for our main conclusions.

\subsection{Evaluation Protocol}
\label{sec:eval_protocol}

Our main experiment tests how well different automated labelers align with human annotators operationalizing the factual-correctness criterion as defined in Section~\ref{sec:constructs}.
For each question–answer pair, automated labels are compared to human-annotated labels to quantify agreement, bias, and error direction. Our primary metric is Cohen's $\kappa$ between each automated labeler and the human labels, reported as a standard chance-corrected agreement measure~\citep{cohen1960coefficient}. We interpret $\kappa$ together with positive-label bias, defined as the automated labeler hallucination rate minus the human labeler hallucination rate ($\hat{p}_{\text{auto}} - \hat{p}_{\text{human}}$), false positive (FP) and false-negative (FN) counts, and FP \% (false positives as a proportion of all errors). To analyze error direction, we compute the false-positive rate $\mathrm{FPR}=FP/(FP+TN)$ and false-negative rate $\mathrm{FNR}=FN/(FN+TP)$ for each label source. These rates distinguish over-labeling from under-labeling, which a single agreement statistic can collapse.

Additionally, we perform a GPT-5-mini ablation experiment, where we report per-generator agreement, bias, and error counts, together with paired prompt-comparison statistics, including point differences in $\kappa$, bootstrap confidence intervals for $\Delta\kappa$, and McNemar's exact two-sided test~\citep{mcnemar1947note}. McNemar's test is appropriate because the prompt variants are evaluated on the same question--answer pairs. It tests whether one prompt corrects more paired errors than it introduces. Full design and analysis details for the factorial ablation are provided in Appendix~\ref{app:factorial_ablation}.

We then use the remaining six LLM judges to assess how far the broad p1$\rightarrow$p2 pattern generalizes across judge families and model capabilities. We also analyze label-source structure using pairwise $\kappa$ agreement matrices and FPR/FNR error-direction plot, with arrows showing movement from p1 to p2 to p3 for multi-prompt judges.

\section{Results}
\label{sec:results}

In this section, we present our results comparing automated labelers to human factual-correctness annotations\footnote{ The code for reproducing our results is shared in our repository: \url{https://github.com/jova486/LPHB}.}. We focus on how prompt criteria, baseline families, and judge types shape errors. 

\subsection{Prompt Criterion Changes Label Behavior}
\label{sec:results_main}

We report the main prompt-ablation results in Table~\ref{tab:results_prompt}. Across the evaluated LLM judges and generator sets, the
faithfulness-style p1 prompt is false-positive-dominant: automated LLM judges often label answers as hallucinated when human
annotators label them factually correct. Interestingly, this pattern is strongest for GPT-5.4 and weakest for Claude Opus 4.7, which are the two frontier judges in our benchmark. A long-form version with one row per judge--generator--prompt combination is provided in Appendix~\ref{app:full_results}, Table~\ref{tab:results_full}.

\begin{table*}[ht!]
  \centering
  \caption{Prompt-ablation results against human labels for each judge--generator pair. Bias is positive when the automated labeler over-labels hallucinations relative to humans. FP/FN reports false positives and false negatives; FP\% is false positives as a proportion of all errors. Best $\kappa$ within each row is shown in bold. All values are computed against \texttt{annotator1\_label}, the complete primary human label set for the 900 question--answer pairs. Gen denotes the answer-generating model: L=Llama-3-8B, G=Gemma-2-9B, and M=Mistral-7B.}
  \label{tab:results_prompt}
  \scriptsize
  \setlength{\tabcolsep}{3pt}
  \renewcommand{\arraystretch}{0.92}
  \begin{tabular}{llrrrrrrrrrrrr}
    \toprule
    & & \multicolumn{4}{c}{p1} & \multicolumn{4}{c}{p2} & \multicolumn{4}{c}{p3} \\
    \cmidrule(lr){3-6}\cmidrule(lr){7-10}\cmidrule(lr){11-14}
    Judge & Gen
    & $\kappa$ & Bias & FP/FN & FP\%
    & $\kappa$ & Bias & FP/FN & FP\%
    & $\kappa$ & Bias & FP/FN & FP\% \\
    \midrule
    \multicolumn{14}{l}{\textit{API judges}} \\
    \midrule
    GPT-5-mini & L & 0.530 & $+$0.223 & 70/3 & 95.9 & \textbf{0.756} & $-$0.020 & 15/21 & 41.7 & 0.744 & $+$0.007 & 20/18 & 52.6 \\
    GPT-5-mini & G & 0.498 & $+$0.193 & 68/10 & 87.2 & \textbf{0.770} & $-$0.033 & 11/21 & 34.4 & 0.759 & $-$0.007 & 16/18 & 47.1 \\
    GPT-5-mini & M & 0.194 & $+$0.383 & 117/2 & 98.3 & 0.626 & $+$0.027 & 32/24 & 57.1 & \textbf{0.646} & $+$0.030 & 31/22 & 58.5 \\
    \addlinespace[1pt]
    GPT-5-nano & L & 0.536 & $+$0.150 & 58/13 & 81.7 & 0.655 & $-$0.017 & 23/28 & 45.1 & \textbf{0.676} & $+$0.000 & 24/24 & 50.0 \\
    GPT-5-nano & G & 0.592 & $+$0.080 & 42/18 & 70.0 & \textbf{0.686} & $-$0.020 & 19/25 & 43.2 & 0.641 & $-$0.027 & 21/29 & 42.0 \\
    GPT-5-nano & M & 0.389 & $+$0.177 & 72/19 & 79.1 & \textbf{0.601} & $-$0.047 & 23/37 & 38.3 & 0.581 & $-$0.057 & 23/40 & 36.5 \\
    \addlinespace[1pt]
    GPT-5.4 & L & 0.306 & $+$0.363 & 110/1 & 99.1 & \textbf{0.779} & $+$0.023 & 20/13 & 60.6 & 0.740 & $+$0.050 & 27/12 & 69.2 \\
    GPT-5.4 & G & 0.240 & $+$0.413 & 127/3 & 97.7 & 0.766 & $-$0.003 & 16/17 & 48.5 & \textbf{0.774} & $+$0.000 & 16/16 & 50.0 \\
    GPT-5.4 & M & 0.021 & $+$0.480 & 144/0 & 100.0 & 0.619 & $+$0.077 & 40/17 & 70.2 & \textbf{0.672} & $+$0.070 & 35/14 & 71.4 \\
    \addlinespace[1pt]
    Claude Opus 4.7 & L & 0.712 & $+$0.037 & 27/16 & 62.8 & \textbf{0.764} & $-$0.003 & 17/18 & 48.6 & 0.738 & $+$0.010 & 21/18 & 53.8 \\
    Claude Opus 4.7 & G & 0.740 & $+$0.067 & 29/9 & 76.3 & \textbf{0.764} & $-$0.023 & 13/20 & 39.4 & 0.737 & $-$0.010 & 17/20 & 45.9 \\
    Claude Opus 4.7 & M & 0.693 & $+$0.033 & 28/18 & 60.9 & \textbf{0.706} & $+$0.013 & 24/20 & 54.5 & 0.673 & $+$0.023 & 28/21 & 57.1 \\
    \midrule
    \multicolumn{14}{l}{\textit{Local judges}} \\
    \midrule
    Gemma-2-9B & L & 0.567 & $+$0.127 & 52/14 & 78.8 & \textbf{0.591} & $-$0.047 & 23/37 & 38.3 & 0.557 & $+$0.020 & 36/30 & 54.5 \\
    Gemma-2-9B & G (self) & 0.463 & $+$0.183 & 69/14 & 83.1 & 0.623 & $-$0.083 & 13/38 & 25.5 & \textbf{0.628} & $-$0.020 & 23/29 & 44.2 \\
    Gemma-2-9B & M & 0.378 & $+$0.090 & 60/33 & 64.5 & \textbf{0.496} & $-$0.153 & 15/61 & 19.7 & 0.455 & $-$0.087 & 28/54 & 34.1 \\
    \addlinespace[1pt]
    Qwen2.5-7B & L & 0.313 & $+$0.280 & 96/12 & 88.9 & \textbf{0.483} & $+$0.023 & 42/35 & 54.5 & 0.472 & $+$0.100 & 55/25 & 68.8 \\
    Qwen2.5-7B & G & 0.376 & $+$0.283 & 93/8 & 92.1 & \textbf{0.607} & $-$0.053 & 19/35 & 35.2 & 0.559 & $+$0.083 & 45/20 & 69.2 \\
    Qwen2.5-7B & M & 0.213 & $+$0.203 & 89/28 & 76.1 & 0.428 & $-$0.060 & 34/52 & 39.5 & \textbf{0.452} & $+$0.047 & 48/34 & 58.5 \\
    \addlinespace[1pt]
    Llama-3-8B & L (self) & \textbf{0.370} & $+$0.077 & 59/36 & 62.1 & 0.303 & $-$0.240 & 13/85 & 13.3 & 0.255 & $-$0.230 & 18/87 & 17.1 \\
    Llama-3-8B & G & 0.227 & $+$0.210 & 92/29 & 76.0 & \textbf{0.470} & $-$0.157 & 11/58 & 15.9 & 0.384 & $-$0.160 & 16/64 & 20.0 \\
    Llama-3-8B & M & \textbf{0.278} & $+$0.073 & 65/43 & 60.2 & 0.218 & $-$0.383 & 2/117 & 1.7 & 0.185 & $-$0.380 & 5/119 & 4.0 \\
    \bottomrule
  \end{tabular}
\end{table*}

Replacing p1 with the factual-correctness p2 prompt generally improves agreement with human-annotated labels and consistently reduces false-positive dominance for nearly all judges. The effect is largest for GPT-5.4 and GPT-5-mini, moderate for GPT-5-nano, Gemma-2-9B, and Qwen2.5-7B, and smallest for Claude Opus 4.7, which is already close to the p2 region under p1. For the primary GPT-5-mini judge, this p1$\to$p2 pattern also appears within individual datasets:  Table~\ref{tab:per_dataset_gpt5mini} in Appendix~\ref{app:gpt_mini_ablation}
shows that the p2 prompt leads to higher agreement than p1 across all datasets and generator models.

The extended p3 prompt does not consistently improve over p2. Across judge families, p2 is usually the best or near-best prompt, indicating that the main empirical change occurs between p1 and p2, after which the effect of correctly specifying the target criterion saturates. For GPT-5-mini, p2 is significantly more often correct than p1 under McNemar's test for all three generators ($p < 0.0001$), and the bootstrap confidence intervals for $\Delta\kappa$ exclude zero in all three comparisons. By contrast, no p2$\to$p3 difference is statistically significant for any generator. Paired prompt-comparison statistics are reported in Appendix~\ref{app:gpt_mini_ablation}, Table~\ref{tab:prompt_comparisons}. 

Because p1 and p2 differ in both labeling criterion and prompt structure, we also test the criterion-alignment interpretation directly with a separate factorial control.
In this 
ablation conducted with GPT-5-mini, we create variants of both p1 and p2 to isolate the effects of labeling criterion and prompt structure. We find that a p1 variant with a prompt structure matching that of p2 results in 
no detectable improvement.
Also, condensing the p2 prompt so that its structure matches that of p1 does not yield worse agreement with the human annotator.
This isolates criterion choice, rather than prompt elaboration, as the main driver of the p1$\to$p2 agreement gain. Full factorial results and prompt definitions are provided in Appendix~\ref{app:factorial_ablation}.

Smaller open-weight local judges are more heterogeneous in their labeling behavior than larger hosted API judges. Gemma-2-9B and Qwen2.5-7B partly reproduce the p1$\to$p2 pattern at lower absolute agreement. Llama-3-8B is the main exception: under p2/p3 it shifts toward a false-negative-dominant profile, especially on Mistral-7B outputs.

\subsection{Lexical and NLI Baselines Fail in Different Directions}
\label{sec:results_lexical}

We report results with ROUGE-L, BERTScore, and NLI baselines against the same human labels under oracle-$\tau$ and 3-fold CV-$\tau$ thresholding in Table~\ref{tab:results_lexical}. We also report METEOR and BARTScore-CNN results in Appendix~\ref{app:new_metrics}, which fall within the range spanned by these baselines. ROUGE-L is false-negative-prone, BERTScore is mixed across generators, and strict NLI is almost entirely false-positive-dominant (FP\%: 98.3--99.4\%). Oracle thresholding improves some scores but does not bring these baselines into the same agreement range as the criterion-aligned API judges.

\begin{table*}[ht!]
  \centering
  \caption{Lexical similarity and NLI baselines against human-annotated factual-correctness labels. Oracle-$\tau$ reports a threshold tuned directly on the evaluation set. CV-$\tau$ reports 3-fold out-of-fold thresholding within each generator. For NLI, best-$\tau$ uses an oracle threshold. Bias is positive when the automated labeler over-labels hallucinations relative to humans. FP\% is false positives as a proportion of all errors. L=Llama-3-8B,
G=Gemma-2-9B, M=Mistral-7B.}
  \label{tab:results_lexical}
  \scriptsize
  \setlength{\tabcolsep}{5pt}
  \renewcommand{\arraystretch}{0.95}
  \begin{tabular}{lllrrrrr}
    \toprule
    Metric & Gen & $\tau$ & $\kappa$ & Bias & FP & FN & FP\% \\
    \midrule
    ROUGE-L (oracle-$\tau$) & L & 0.06 & 0.403 & $-$0.183 & 15 & 70 & 17.6 \\
    ROUGE-L (oracle-$\tau$) & G & 0.07 & 0.483 & $-$0.057 & 27 & 44 & 38.0 \\
    ROUGE-L (oracle-$\tau$) & M & 0.03 & 0.412 & $-$0.257 & 6 & 83 & 6.7 \\
    ROUGE-L (CV-$\tau$) & L & 0.07 [0.06, 0.10] & 0.353 & $-$0.143 & 25 & 68 & 26.9 \\
    ROUGE-L (CV-$\tau$) & G & 0.07 [0.06, 0.07] & 0.473 & $-$0.067 & 26 & 46 & 36.1 \\
    ROUGE-L (CV-$\tau$) & M & 0.03 [0.03, 0.04] & 0.385 & $-$0.230 & 12 & 81 & 12.9 \\
    \midrule
    BERTScore (oracle-$\tau$) & L & 0.85 & 0.387 & $-$0.047 & 38 & 52 & 42.2 \\
    BERTScore (oracle-$\tau$) & G & 0.85 & 0.486 & $+$0.117 & 56 & 21 & 72.7 \\
    BERTScore (oracle-$\tau$) & M & 0.84 & 0.415 & $-$0.067 & 34 & 54 & 38.6 \\
    BERTScore (CV-$\tau$) & L & 0.85 [0.85, 0.86] & 0.318 & $-$0.003 & 50 & 51 & 49.5 \\
    BERTScore (CV-$\tau$) & G & 0.85 [0.85, 0.85] & 0.486 & $+$0.117 & 56 & 21 & 72.7 \\
    BERTScore (CV-$\tau$) & M & 0.84 [0.84, 0.85] & 0.394 & $-$0.037 & 40 & 51 & 44.0 \\
    \midrule
    NLI (T5-11B) strict & L & 0.50 & 0.053 & $+$0.513 & 155 & 1 & 99.4 \\
    NLI (T5-11B) strict & G & 0.50 & 0.025 & $+$0.573 & 175 & 3 & 98.3 \\
    NLI (T5-11B) strict & M & 0.50 & 0.049 & $+$0.453 & 138 & 2 & 98.6 \\
    NLI (T5-11B) best-$\tau$ & L & 0.01 & 0.380 & $+$0.100 & 62 & 32 & 66.0 \\
    NLI (T5-11B) best-$\tau$ & G & 0.01 & 0.217 & $+$0.200 & 91 & 31 & 74.6 \\
    NLI (T5-11B) best-$\tau$ & M & 0.01 & 0.395 & $-$0.077 & 34 & 57 & 37.4 \\
    \bottomrule
  \end{tabular}
\end{table*}

Cross-validated thresholds are close to oracle thresholds within each generator set, suggesting that poor performance is not mainly due to unstable within-generator threshold selection. The NLI best-$\tau$ results are also informative: the oracle threshold collapses to $\tau=0.01$ for all generators, yet agreement remains modest. The main remaining variation is across generators, especially for ROUGE-L, whose optimal threshold is lower for longer Mistral-7B answers. We return to the construct-level implication of these error profiles in
Section~\ref{sec:discussion}.

\subsection{Error Direction Separates Label Sources}
\label{sec:error_direction_structure}

\begin{figure}[t!]
  \centering

  \includegraphics[width=0.8\columnwidth]{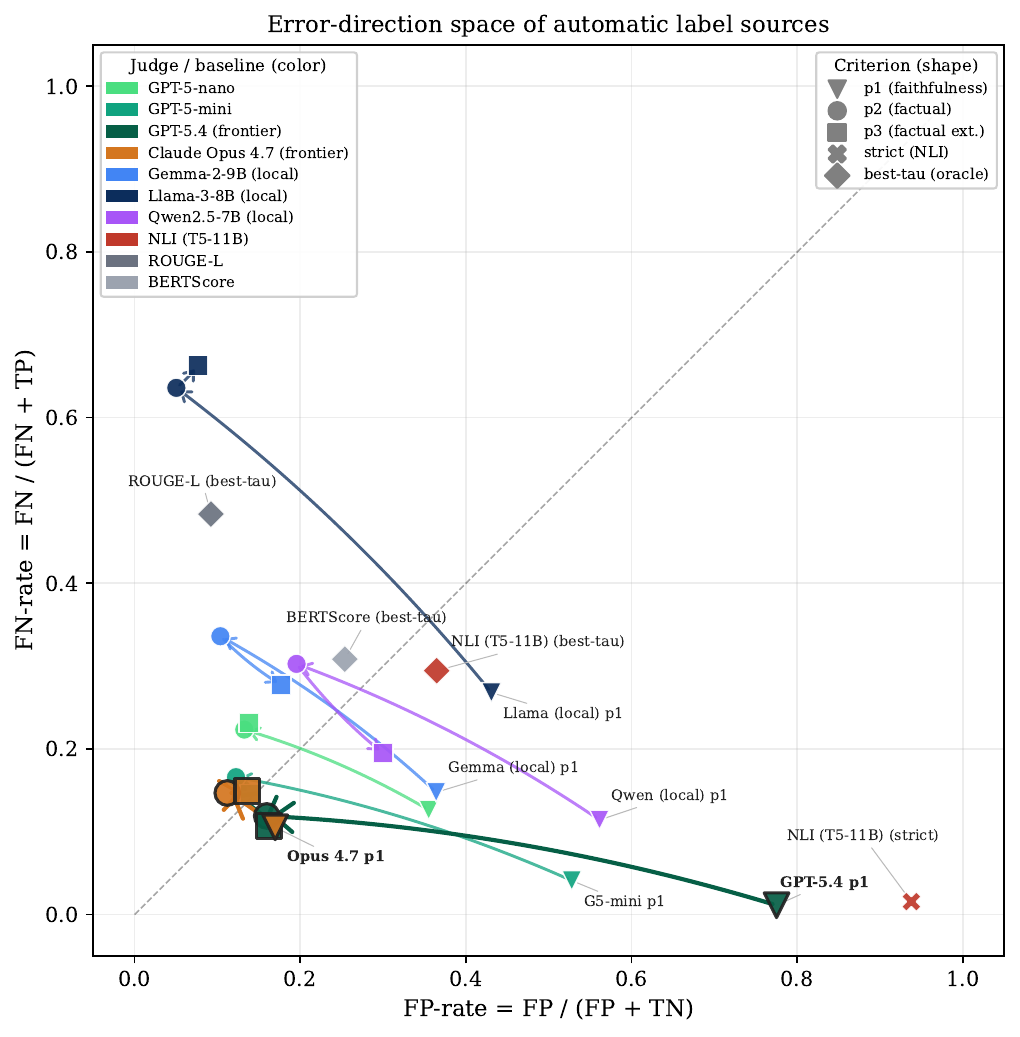}
  \caption{Error-direction space of automated label sources. Each point shows macro-averaged false-positive and false-negative rates over the three generator sets. The main p1$\to$p2 movement is a reduction in false positives, while local judges and non-LLM
  baselines occupy distinct regions.}
  \label{fig:error_direction_space}
\end{figure}

Pairwise inter-method agreement matrices (see Appendix~\ref{app:per_generator_matrices}) show that criterion-aligned judges form a coherent agreement block, while lexical metrics, strict NLI, and faithfulness-style p1 sources show weaker agreement with all other label sources. The geometry of the distinct regimes of the label sources in false-positive / false-negative space is illustrated in Figure~\ref{fig:error_direction_space}. The main p1$\to$p2 movement for API judges is leftward, corresponding primarily to fewer false positives. GPT-5.4 p1 is the most extreme false-positive-dominant LLM judge, whereas for Claude Opus 4.7, the p1 region is much closer to the p2 region. Local judges occupy more varied positions, with the p2/p3 regions for Llama-3-8B moving into a false-negative-dominant region from the respective p1 region. The non-LLM baselines also separate by error direction: strict NLI is false-positive-dominant, ROUGE-L is mostly false-negative-prone, and BERTScore is mixed.

\section{Discussion}
\label{sec:discussion}

In this section, we interpret our empirical findings and discuss limitations. We also outline some practical implications for building and validating hallucination benchmarks.

\subsection{Faithfulness Bias as Measurement Error}
\label{sec:discussion_faithfulness}

The central result of this study is not only that some automatically generated labels disagree with human labels, but that the disagreement is directional. When factual correctness is the intended target, faithfulness-style label sources tend to over-label factually correct elaborations as hallucinations. A paired elaboration-removal intervention supports this interpretation: removing elaborative material from p1-evaluated answers reduced the false-positive rate from 52.9\% to 23.2\% (45 FP→TN vs. 4 TN→FP transitions; see Appendix~\ref{app:elaboration_removal}).

This matters because both criteria can be represented with the same binary label and the same term, ``hallucination,'' while measuring different constructs. Our aim is not to argue that factual correctness is always the correct target. In evidence-grounded settings such as RAG or summarization, source faithfulness may be the intended construct. The problem arises when a benchmark intends to evaluate factual correctness but uses a label source that operationalizes reference faithfulness.

\subsection{The Main Intervention Is Criterion Alignment}
\label{sec:discussion_criterion}

The p1$\to$p2 ablation suggests that the main intervention is criterion alignment, not prompt elaboration. The p3 prompt adds more explicit decision structure but does not consistently improve over p2. The important rule is the distinction between absence and contradiction: information absent from a short reference should not automatically be treated as hallucination. The labeling criterion vs. prompt structure factorial ablation (see Appendix~\ref{app:factorial_ablation}) supports this interpretation directly: holding the factual-correctness criterion fixed while adding decision structure produces no detectable change in agreement, whereas changing the criterion improves agreement at both levels of structure.

This distinction is small at the level of prompt wording but substantial
for what the resulting binary labels measure. Without it, an automated LLM judge may silently convert a factual-correctness task into a reference-coverage task. Making the distinction explicit can therefore move the judge closer to the human factual-correctness labels without changing the dataset, the generated answers, or the binary labeling format.

\subsection{Judge Capability Is Not Enough}
\label{sec:discussion_judges}

The altered behavior of frontier judges under different prompting schemes shows that criterion alignment is not determined by capability alone. GPT-5.4 and Claude Opus 4.7 behave very differently under the same nominal p1 prompt, but move toward a more similar region under p2. We use this contrast not as a model-ranking claim, but as evidence that the judge's default interpretation of an ambiguous criterion is itself an evaluation choice. Our experiments do not identify the mechanism underlying these cross-model differences, which we leave for future work.

Local judges further show that structured prompts are not sufficient by themselves. Gemma-2-9B and Qwen2.5-7B partly reproduce the criterion-alignment pattern at lower absolute agreement, whereas Llama-3-8B shifts toward a false-negative-dominant profile under p2/p3. This suggests a robustness limit of smaller models on realistic QA outputs rather than a failure to understand the abstract rule.

The same caution applies to non-LLM label sources. Lexical similarity metrics and strict NLI do not merely require better threshold calibration; they impose distinct operational criteria. In particular, answer--reference similarity can penalize elaboration even when the additional content is factually correct. Similarly, strict entailment can label correct information absent from the reference as non-entailed, and therefore as hallucinated under a reference-faithfulness criterion. 

\subsection{Implications and Limitations}
\label{sec:discussion_implications_limitations}

These results suggest three practical recommendations for open-domain QA hallucination benchmarks. First, benchmark papers should specify whether labels target reference faithfulness or factual correctness. Second, when factual correctness is the intended target, judge prompts should explicitly distinguish contradiction from absence and allow correct elaboration for the models being evaluated. Third, automated label sources should be validated on a small human-labeled set before being used as ground truth, because a judge's capability alone does not guarantee criterion alignment. This validation must target the intended criterion: agreement among raters applying a reference-faithfulness criterion does not by itself establish that the labels measure factual correctness.

Our results also bear on how existing benchmarks should be read. As surveyed in Section~\ref{sec:related}, many benchmark evaluations derive labels from the relationship between an answer and a supplied reference, so their conclusions are conditional on the chosen labeling rule. Results based on reference similarity should therefore be interpreted as performance under that specific operational criterion, rather than as general evidence of hallucination-detection or factual-correctness performance.

This study was limited to English open-domain QA with short references and three relatively small open-weight generators. The benchmark was constructed to compare label sources on a fixed human-labeled set produced under a limited annotation budget;
a larger sample would narrow the intervals reported here. Validation across further task families, longer answers, domains, and generators remains future work.

Our error counts are defined relative to human labels collected under a single criterion: the false positives reported for faithfulness-style sources are errors with respect to $y_{\text{fact}}$, not in an absolute sense, and comparison against human annotations collected under a reference-faithfulness criterion is left to future work. The human labels themselves are not error-free: the main experiments use the A1 labels rather than consensus-adjudicated labels, and we did not log how often annotators consulted external sources, which limits our ability to quantify how often the short references were insufficient.
It is also worth noting that answer-level binary labels cannot capture error severity or partial correctness.

LLM-based generation and judging may also vary across random seeds, implementations, provider-side model versions, and inference configurations. The reported values should therefore be read as conditional on the fixed generated-answer set, judge versions, prompts, and inference setups evaluated here. The full results in Table~\ref{tab:results_full} in Appendix~\ref{app:full_results} also include self-evaluation conditions for Gemma-2-9B and Llama-3-8B, which are marked explicitly. These rows are reported for completeness but are not used as evidence that self-evaluation is unbiased. To ensure that all judges are able to follow the p2 factual-correctness decision rule, we conducted a six-branch sanity check on simple controlled cases. All seven judges passed this check (Appendix~\ref{app:sanity}, Table~\ref{tab:sanity_check}). 

\section{Conclusion}
\label{sec:conclusion}

This paper examined a basic but often implicit assumption in automated hallucination evaluation: that different label sources using the same binary label measure the same construct. In short-reference open-domain QA, this assumption can fail. A faithfulness-style label source may penalize answers for adding information not contained in a short reference, even when the added information is factually correct. When the intended benchmark target is factual correctness, this creates a systematic form of measurement error.

Across seven LLM judges, three generator families, lexical similarity metrics, and a reference-entailment NLI baseline, we find that label-source choice changes both agreement with human labels and the direction of errors. Faithfulness-style prompts are often false-positive-dominant, while a factual-correctness prompt that explicitly distinguishes absence from contradiction substantially reduces this bias for most judges. 
The effect is not reducible to judge capability: frontier judges can behave very differently under the same nominal prompt, and local judges show additional robustness limits. Non-LLM baselines also fail in distinct directions, with ROUGE-L, BERTScore, and strict NLI occupying different regions of error-direction space.

The broader implication is that hallucination labels are not interchangeable implementation details. They are part of the benchmark design. Benchmarks should state whether they target reference faithfulness or factual correctness, and automated label sources should be validated against that intended criterion. For short-reference QA, a minimal but consequential rule is that information absent from the reference should not be treated as hallucination unless it is contradicted or factually false. Making this criterion explicit can turn automated hallucination evaluation from a surface label-matching exercise into a more valid measurement of the intended construct.

\section*{Acknowledgments and Disclosures}

This work was partly supported by Business Finland under
grant agreement 23004 ELFMo of the ITEA4 programme.
This funding supported three of the authors. The authors declare no competing interests.

\bibliographystyle{plainnat}
\bibliography{labeling_submission_ready}

\appendix

\section{Annotation Protocol and Agreement}
\label{app:annotation}

In this appendix, we describe the annotation protocol used in the human-annotated benchmark.

\paragraph{Human Annotation Criterion}
Annotators apply the factual-correctness criterion defined in Section~\ref{sec:constructs}. The label is assigned at the answer level. An answer is labeled as a hallucination ($y=1$) if it contains either (i) a claim contradicted by the reference answer, or (ii) a specific verifiable factual claim, such as a date, name, number, place, organization, entity, or other identifiable detail, that is judged factually incorrect based on the reference answer, or, when the reference is insufficient, external factual adjudication. 

An answer is labeled non-hallucinated ($y=0$) if its verifiable specific claims are factually correct. Information absent from the reference is not by itself sufficient for a hallucination label: if an added detail is verifiably true, it is not penalized under $y_{\text{fact}}$. Conversely, a correct core answer does not override a fabricated or factually incorrect specific detail elsewhere in the response.

\paragraph{Annotation Procedure}
For each question--answer pair, annotators follow this workflow:
\begin{enumerate}
    \item Read the question.
    \item Identify the specific verifiable claims and other identifiable
          factual details in the model answer.
    \item Read the reference answer(s).
    \item For each specific claim, determine whether it is
          (a) contradicted by the reference, or
          (b) unresolved by the reference, but factually false under
          external adjudication.
    \item Assign label 1 if any claim fails step 4; otherwise, assign
          label 0.
\end{enumerate}

\paragraph{External Factual Adjudication}
When the reference answer is insufficient to resolve a specific verifiable claim, annotators consult external evidence. In this study, external adjudication relied on public sources such as Wikipedia and general web searches. AI-assisted fact-checking could be used as a support tool, but not as the sole basis for assigning a label. The frequency of external adjudication was not systematically logged. Unsupported but true elaborations are not penalized.

\paragraph{Operational Scope}
This protocol is reference-first, but not strictly reference-only. Short reference answers included in the QA datasets are treated as primary evidence rather than as a complete specification of all true answer content. Accordingly, absence from the reference is not sufficient for a hallucination label. The operational target is answer-level factual correctness for specific verifiable claims, rather than strict reference faithfulness.

The protocol distinguishes hallucination labels from general answer quality. A vague, incomplete, or non-committal answer is not labeled as a hallucination unless it contradicts the reference or contains a specific, verifiable false claim. Likewise, an answer may be factually correct under this protocol, even if it contains correct information not present in the short reference.

\paragraph{Boundary Cases}
The following examples illustrate the distinction between reference faithfulness and factual correctness:
\begin{itemize}
    \item \textit{Q:What is the capital of France?}
    \begin{itemize}
        \item \textit{Ref: Paris. Answer: ``Paris, the capital of France.''}
          $\rightarrow$ \textbf{0}\\
          The added phrase is absent from the reference, but verifiably
          correct, so the answer is not labeled hallucination under
          $y_{\text{fact}}$.

        \item \textit{Ref: Paris. Answer: ``Paris, founded in 1887.''}
          $\rightarrow$ \textbf{1}\\
          The core answer is correct, but the added founding date is a specific verifiable claim that is factually incorrect under external adjudication.

        \item \textit{Ref: Paris. Answer: ``Lyon.''}
          $\rightarrow$ \textbf{1}\\
          The answer directly contradicts the reference.

        \item \textit{Ref: Paris. Answer: ``I am not sure.''}
          $\rightarrow$ \textbf{0}\\
          The answer is non-committal and may be unhelpful, but it does
          not assert a contradictory or fabricated specific claim.
    \end{itemize}

    \item \textit{Q: What color is the sky on a clear day?}
    
    \begin{itemize}
        \item \textit{Ref: Blue.
          Answer: ``Blue, specifically a shade known as cerulean.''}
          $\rightarrow$ \textbf{0}\\
          The elaboration is absent from the reference but verifiably
          correct, so it is not penalized under $y_{\text{fact}}$.
    \end{itemize}
\end{itemize}

\paragraph{Primary Benchmark Labels}
All 900 benchmark items are labeled by a primary annotator (A1) under the protocol above. These A1 labels serve as the benchmark labels for all downstream evaluation reported in the paper.

\paragraph{Independent Reliability Annotation}
To estimate annotation reliability, a second annotator (A2) independently labeled a 300-item subset of the benchmark spanning all three datasets and all three generator models. The A2 labels are used only for reliability estimation and to support quality control for the annotation protocol, quantifying the stability of the protocol across datasets and generator models. The main experiments do not replace A1 labels with consensus or adjudicated labels. Furthermore, this inter-annotator analysis is not to be taken as a claim that the benchmark's labels are error-free.

\paragraph{Inter-Annotator Agreement}
Table~\ref{tab:iaa} reports independent agreement on the 300-item subset. No disagreement-based corrections were applied to either annotator before
computing these values. Overall agreement is strong, with Cohen's $\kappa=0.808$ and raw agreement of 90.3\% (271/300). The 95\% confidence
interval for $\kappa$ was computed with 1,000 question-level bootstrap resamples~\citep{efron1993bootstrap}, keeping the three generator responses to
each sampled question together. The resulting overall interval is $[0.729, 0.879]$. Dataset-level point estimates are highest for HotpotQA
($\kappa=0.832$) and lower for TruthfulQA ($\kappa=0.690$), although the
dataset-specific confidence intervals are wide and these differences should
be interpreted descriptively.

\begin{table}[t]
\centering
\caption{Independent inter-annotator agreement on the 300-item subset.
Confidence intervals are 95\% question-level bootstrap intervals for Cohen's
$\kappa$ with 1,000 resamples, keeping the three generator responses to each
question together.}
\label{tab:iaa}
\small
\begin{tabular}{lrrrrrr}
\toprule
Dataset & $n$ & Disagr. & Agree. & $\kappa$ & CI low & CI high \\
\midrule
TriviaQA   &  99 &  9 & 0.909 & 0.796 & 0.618 & 0.935 \\
TruthfulQA &  99 & 14 & 0.859 & 0.690 & 0.544 & 0.816 \\
HotpotQA   & 102 &  6 & 0.941 & 0.832 & 0.642 & 0.968 \\
\midrule
Overall    & 300 & 29 & 0.903 & 0.808 & 0.729 & 0.879 \\
\bottomrule
\end{tabular}
\end{table}

\section{Judge Prompts}
\label{app:judge_prompts}

This appendix provides the three main prompt definitions used in the cross-judge experiments. The two auxiliary prompts used only in the GPT-5-mini criterion-by-structure factorial ablation (p1s and p2t) are provided with that experiment in Appendix~\ref{app:factorial_ablation}.

\paragraph{p1 (faithfulness baseline)}
This prompting strategy follows a reference-grounded criterion. It labels an answer as a hallucination if a claim is
unsupported by or contradicted by the reference. This prompt operationalizes the faithfulness-style criterion used in the main prompt-ablation experiments.

\begin{lstlisting}[style=promptstyle]
You are a hallucination auditor for open-domain QA.

Your task: decide if a model answer contains a hallucination by comparing
it against the reference answer.

DEFINITION
A hallucination is any claim that is not supported by, or contradicted by,
the reference answer.

Use the reference answer as ground truth. Do not use external knowledge.

OUTPUT FORMAT (STRICT JSON ONLY)
Return ONLY valid JSON with exactly these two fields:
{"label": 0 or 1, "confidence": <float between 0 and 1>}

Rules:
- Do not include markdown, code fences, or extra text.
- Do not include a reasoning field.
- confidence must be a numeric value (not a string).
\end{lstlisting}

\paragraph{p2 (factual-correctness decision tree)}
This is a structured prompt with explicit rules separating contradiction from mere
absence: \emph{``Absent from reference'' is not the same as
``contradicted by reference.''} A hallucination requires either explicit
contradiction or a verifiable false specific detail --- directly
operationalizing the annotation protocol described in Appendix~\ref{app:annotation}.
Vague or non-committal answers are not hallucinations unless they
contradict the reference or include a factually false specific claim.
 
\begin{lstlisting}[style=promptstyle]
You are a hallucination auditor for open-domain QA.

Your task: decide if a model answer contains a hallucination.

DEFINITION
A hallucination is a claim that is EXPLICITLY CONTRADICTED by the reference
answer, internally self-contradictory, or factually false at the level of a
specific verifiable detail (such as a name, date, number, place, entity, or
other concrete factual attribute).

CRITICAL RULES
1. "Absent from reference" is NOT the same as "contradicted by reference".
   Additional correct details not in the reference are NOT hallucinations.
2. Wrong-but-vague is NOT a hallucination.
   Vague, incomplete, or non-committal answers are incorrect, not
   hallucinations, unless they explicitly contradict the reference or include
   a specific verifiable false detail.
3. Focus on factual error, not mere non-overlap.
   Do not label an answer as a hallucination only because it contains content
   that does not appear in the reference.

DECISION TREE
Q1. Does the answer match the reference or a clear paraphrase?
    YES -> Q2. Does it add any claim EXPLICITLY CONTRADICTED by the reference or 
              factually false at the level of a specific verifiable detail?
              NO  -> LABEL=0
              YES -> LABEL=1
    NO  -> Q3. Does the core answer conflict with the reference (different
              entity/value/fact)?
              YES -> LABEL=1
              NO  -> Q4. Is the answer a refusal or expression of uncertainty?
                        YES -> LABEL=0
                        NO  -> Q5. Does it include a specific verifiable detail
                                  that is factually false?
                                  YES -> LABEL=1
                                  NO  -> LABEL=0

OUTPUT FORMAT (STRICT JSON ONLY)
Return ONLY valid JSON with exactly these two fields:
{"label": 0 or 1, "confidence": <float between 0 and 1>}

Rules:
- Do not include markdown, code fences, or extra text.
- Do not include a reasoning field.
- confidence must be a numeric value (not a string).
\end{lstlisting}

\paragraph{p3 (extended factual-correctness decision tree)}
This prompting strategy extends p2 with an explicit branch for non-committal references (e.g., ``unknown,'' refusal) and additional edge-case handling. The criterion remains factual correctness; p3 is treated as an exploratory edge-case variant that adds a more refined decision structure rather than a different construct.

\begin{lstlisting}[style=promptstyle]
You are a hallucination auditor for open-domain QA.

Your task: decide if a model answer contains a hallucination.

DEFINITION
A hallucination is a claim that is EXPLICITLY CONTRADICTED by the reference
answer, internally self-contradictory, or factually false at the level of a
specific verifiable detail (such as a name, date, number, place, entity, or
other concrete factual attribute).

CRITICAL RULES
1. "Absent from reference" is NOT the same as "contradicted by reference".
   Additional correct details not in the reference are NOT hallucinations.
2. Wrong-but-vague is NOT a hallucination.
   Vague or incomplete answers are incorrect, not hallucinations, unless
   they explicitly contradict the reference or include a specific verifiable
   false detail.
3. Core mismatch = hallucination.
   If the core answer conflicts with the reference (different entity/value/fact),
   label=1 even without an additional false specific detail.

SPECIAL CASE -- non-committal references:
If the reference is an explicit refusal or non-answer (e.g., "I have no comment",
"Unknown", "Cannot be determined"), then any confident specific claim from the
model is a hallucination. If the reference explains a dependency ("It depends
on X because..."), evaluate normally.

DECISION TREE
Q1. Does the answer match the reference or a clear paraphrase?
    YES -> Q2. Does it add any claim EXPLICITLY CONTRADICTED by the reference or 
              factually false at the level of a specific verifiable detail?
              NO  -> LABEL=0
              YES -> LABEL=1
    NO  -> Q3. Is the reference non-committal (refusal / unknown / cannot be
              determined)?
              YES -> Is the model answer a confident specific claim?
                    YES -> LABEL=1
                    NO  -> LABEL=0
              NO  -> Q4. Does the core answer conflict with the reference?
                        YES -> LABEL=1
                        NO  -> Q5. Is the answer a refusal/uncertainty?
                                  YES -> LABEL=0
                                  NO  -> Q6. Does it include a specific
                                            verifiable detail that is factually
                                            false?
                                            YES -> LABEL=1
                                            NO  -> LABEL=0

OUTPUT FORMAT (STRICT JSON ONLY)
Return ONLY valid JSON with exactly these two fields:
{"label": 0 or 1, "confidence": <float between 0 and 1>}

Rules:
- Do not include markdown, code fences, or extra text.
- Do not include a reasoning field.
- confidence must be a numeric value (not a string).
\end{lstlisting}

\section{Experiment Setup}

In this appendix, we describe details about our experiment setup.

\subsection{Answer Generation Setup}
\label{app:generation}

Answers are generated using the HuggingFace \texttt{transformers}\footnote{\url{https://huggingface.co/docs/transformers/index}} library with each model's default instruction-tuned chat template (\texttt{apply\_chat\_template}). The system prompt is \texttt{"You are a helpful assistant."} for all three generators. Generation uses nucleus sampling with temperature \(T = 0.7\), top-\(p = 0.95\), and a base random seed of 42. The \texttt{max\_new\_tokens} limit is 128 for Gemma-2-9B and Mistral-7B and 96 for Llama-3-8B. Answers that reach this limit without an end-of-sequence token end mid-text; they were labeled in this truncated form (Appendix~\ref{app:truncation}). Model identifiers are \texttt{meta-llama/Meta-Llama-3-8B-Instruct}, \texttt{mistralai/Mistral-7B-Instruct-v0.3}, and \texttt{google/gemma-2-9b-it}.

\subsection{Judge Evaluation Setup}
\label{app:judge_setup}

Judges are prompted with the p1, p2, and p3 templates described in Appendix~\ref{app:judge_prompts} and are required to return strict JSON outputs with a binary \texttt{label} and numeric \texttt{confidence} (on the interval [0,1]). Responses that cannot be parsed as valid JSON are excluded only if automatic repair is impossible. We summarize the judges used, their providers, and their role in our experiments in Table~\ref{tab:judge_setup}.

\begin{table}[ht!]
\centering
\caption{Judge model identifiers and evaluation roles.}
\label{tab:judge_setup}
\small
\begin{tabular}{lll}
\toprule
Judge & Provider / deployment & Role in experiments \\
\midrule
GPT-5-mini & OpenAI API & Primary ablation judge \\
GPT-5-nano & OpenAI API & Weaker API judge \\
GPT-5.4 & OpenAI API & Frontier OpenAI judge \\
Claude Opus 4.7 & Anthropic API & Frontier Anthropic judge \\
Gemma-2-9B-it & Local open-weight & Local judge; includes one self condition \\
Llama-3-8B-Instruct & Local open-weight & Local judge; includes one self condition \\
Qwen2.5-7B-Instruct & Local open-weight & Independent local judge \\
\bottomrule
\end{tabular}
\end{table}

\subsection{NLI Fine-tuning Datasets}\label{app:nli-fine-tuning-datasets}

The choice to use T5-11B as a literature-anchored entailment baseline alongside the lexical metrics was motivated by \citet{Yin_Lid}, who use a T5-based NLI approach
following \citet{honovich2022true} in the reference-as-premise,
answer-as-hypothesis direction. We therefore report conclusions specific to
this T5-11B instantiation and do not claim that it is optimal or representative
of NLI systems in general.
As briefly discussed in Section~\ref{sec:label_sources}, as our reference-entailment NLI baseline, we used a fine-tuned T5-11B model similarly to TRUE~\citep{honovich2022true}, but instead of using the version fine-tuned with the adversarial ANLI dataset~\citep{nie2019adversarial}, we used a version that was fine-tuned with a larger collection of datasets~\footnote{This version is available at \url{https://huggingface.co/google/t5_xxl_true_nli_mixture}}. These datasets are (SNLI~\citep{bowman2015}, MNLI~\citep{williams-etal-2018-broad}, Fever~\citep{thorne-etal-2018-fever}, Scitail~\citep{10.5555/3504035.3504671}, PAWS~\citep{zhang2019pawsparaphraseadversariesword}, VitaminC~\citep{schuster2021vitamincrobustfact}).

\subsection{Truncated Answers}
\label{app:truncation}
 
Of the 900 answers, 258 (28.7\%) reached the generation length limit without an end-of-sequence token and ended mid-text, most of them in TruthfulQA (Table~\ref{tab:truncation_counts}). They were labeled in this truncated form, which is also the form in the dataset. The dataset release marks them in the \texttt{truncated} column and records each answer's length in \texttt{answer\_tokens}.
 
Truncated answers reach their generator-specific maximum lengths and lie at the long end of the answer-length distribution. Agreement between automated
label sources and the primary human annotator is lower on them than on
complete answers, except for strict NLI, and the difference is smaller for the second human annotator (Table~\ref{tab:truncation_kappa}). Among complete
answers, agreement also decreases with answer length: the three p2 judges with
the highest overall agreement (GPT-5-mini, GPT-5.4 and Claude Opus 4.7) disagree with
A1 on 4--5\% of answers of at most 32 tokens, on 18--19\% of complete answers
of 96 tokens or more, and on 20--23\% of truncated answers. The same monotonic
increase in disagreement across answer-length bins is present for all seven p2
judges. Thus, the lower agreement on truncated answers is not specific to
truncation but follows a general decline in agreement as answer length
increases. The p1$\rightarrow$p2 criterion effect for GPT-5-mini and GPT-5.4
is present on both complete and truncated answers.

\begin{table}[ht]
\centering
\caption{Answers truncated at the generation length limit (truncated / all answers), by dataset and generator.}
\label{tab:truncation_counts}
\small
\begin{tabular}{lcccc}
\hline
Dataset & Llama-3-8B & Gemma-2-9B & Mistral-7B & Total \\
\hline
TriviaQA & 5/100 & 4/100 & 18/100 & 27/300 \\
HotpotQA & 9/101 & 10/101 & 20/101 & 39/303 \\
TruthfulQA & 75/99 & 69/99 & 48/99 & 192/297 \\
\hline
Total & 89/300 & 83/300 & 86/300 & 258/900 \\
\hline
\end{tabular}
\end{table}

\begin{table}[ht]
\centering
\caption{Cohen's $\kappa$ with the primary human annotator (A1) on complete ($n=642$) and truncated ($n=258$) answers, pooled over generators, with 95\% question-cluster bootstrap intervals (2000 resamples of the 300 questions). The last row is the second human annotator (A2) on the double-annotated subset ($n=213$ complete, $n=87$ truncated).}
\label{tab:truncation_kappa}
\small
\begin{tabular}{llccc}
\hline
Label source & Prompt & Complete & Truncated & Difference \\
\hline
GPT-5-mini & p1 & 0.51 [0.45, 0.57] & 0.25 [0.17, 0.33] & 0.26 \\
 & p2 & 0.80 [0.74, 0.85] & 0.53 [0.41, 0.64] & 0.27 \\
 & p3 & 0.80 [0.74, 0.85] & 0.52 [0.40, 0.62] & 0.28 \\
GPT-5-nano & p1 & 0.62 [0.56, 0.69] & 0.27 [0.17, 0.37] & 0.35 \\
 & p2 & 0.74 [0.67, 0.79] & 0.43 [0.31, 0.54] & 0.31 \\
 & p3 & 0.74 [0.68, 0.80] & 0.37 [0.25, 0.48] & 0.37 \\
GPT-5.4 & p1 & 0.30 [0.24, 0.35] & 0.03 [0.01, 0.06] & 0.26 \\
 & p2 & 0.79 [0.73, 0.84] & 0.56 [0.44, 0.67] & 0.23 \\
 & p3 & 0.78 [0.72, 0.84] & 0.60 [0.50, 0.69] & 0.18 \\
Claude Opus 4.7 & p1 & 0.80 [0.75, 0.85] & 0.51 [0.40, 0.62] & 0.29 \\
 & p2 & 0.81 [0.76, 0.86] & 0.58 [0.47, 0.69] & 0.23 \\
 & p3 & 0.80 [0.74, 0.85] & 0.53 [0.41, 0.63] & 0.27 \\
Gemma-2-9B & p1 & 0.63 [0.57, 0.70] & 0.13 [0.04, 0.23] & 0.50 \\
 & p2 & 0.64 [0.57, 0.71] & 0.38 [0.25, 0.50] & 0.26 \\
 & p3 & 0.63 [0.56, 0.70] & 0.33 [0.21, 0.45] & 0.30 \\
Qwen2.5-7B & p1 & 0.44 [0.36, 0.51] & 0.03 [$-0.04$, 0.12] & 0.40 \\
 & p2 & 0.58 [0.51, 0.65] & 0.32 [0.20, 0.44] & 0.27 \\
 & p3 & 0.61 [0.54, 0.69] & 0.22 [0.11, 0.34] & 0.39 \\
Llama-3-8B & p1 & 0.41 [0.33, 0.49] & 0.05 [$-0.05$, 0.15] & 0.36 \\
 & p2 & 0.35 [0.28, 0.42] & 0.22 [0.12, 0.33] & 0.13 \\
 & p3 & 0.29 [0.22, 0.36] & 0.18 [0.07, 0.29] & 0.11 \\
\hline
ROUGE-L (CV-$\tau$) & -- & 0.45 [0.37, 0.53] & 0.25 [0.12, 0.37] & 0.21 \\
BERTScore (CV-$\tau$) & -- & 0.45 [0.37, 0.54] & 0.24 [0.11, 0.38] & 0.21 \\
NLI (strict) & -- & 0.03 [0.01, 0.06] & 0.05 [0.00, 0.11] & $-0.02$ \\
\hline
A2 (human) & -- & 0.83 [0.73, 0.91] & 0.75 [0.61, 0.86] & 0.09 \\
\hline
\end{tabular}
\end{table}

\section{Ablation Studies}

In this appendix, we present additional experiments and their results. These are used to further clarify our claims and to offer additional insights.

\subsection{Six-Branch Sanity Check}
\label{app:sanity}

We conducted a simple six-branch sanity check to ensure that all LLM judges used in our experiments were able to follow the instructions given in the p2 factual-correctness prompt. The six branches are: (1) exact correct answer, (2) correct answer with true
elaboration absent from the reference, (3) correct core answer with a false
added specific detail, (4) wrong answer with fabricated specific detail,
(5) core answer mismatch, and (6) refusal or uncertainty expression. The results of this sanity check are shown in Table~\ref{tab:sanity_check}. All seven judges pass the diagnostic cases, including the key branch where a correct answer adds true information absent from the reference. This indicates that the prompt is understandable in simple controlled cases. The main benchmark results, therefore, reflect differences in robustness on realistic QA outputs, not a complete inability to follow the basic decision rule.

\begin{table}[ht!]
  \centering
  \caption{Six-branch sanity check for the p2 factual-correctness prompt. The check tests basic prompt-following behavior on simple diagnostic cases and is not used to estimate benchmark performance.}
  \label{tab:sanity_check}
  \small
  \begin{tabular}{llr}
    \toprule
    Judge & Type & Score \\
    \midrule
    GPT-5-mini & API & 6/6 \\
    GPT-5-nano & API & 6/6 \\
    GPT-5.4 & API & 6/6 \\
    Claude Opus 4.7 & API & 6/6 \\
    Gemma-2-9B & Local & 6/6 \\
    Llama-3-8B & Local & 6/6 \\
    Qwen2.5-7B & Local & 6/6 \\
    \bottomrule
  \end{tabular}
\end{table}

\subsection{GPT-5-mini ablations}
\label{app:gpt_mini_ablation}

In this appendix, we describe more detailed experiments limited to GPT-5-mini.

\subsubsection{Paired Prompt Comparisons}
\label{app:paired_prompt_comparisons}

We use paired prompt comparisons to separate two questions. First, we measure how well each prompt variant agrees with the human labels using Cohen's $\kappa$, which corrects nominal-label agreement for chance. Second, because the prompt variants are evaluated on the same generated answers, we compare their item-level correctness with an exact two-sided McNemar's test over discordant correctness outcomes. Thus, $\Delta\kappa$ summarizes the change in chance-corrected agreement with the human labels, while the McNemar's test asks whether one prompt is more often correct than the other on the same examples. We report nonparametric bootstrap intervals for $\Delta\kappa$ to quantify the sampling uncertainty of the agreement difference. The results are reported in Table~\ref{tab:prompt_comparisons}.

The results show that switching from p1 (faithfulness) to p2 (factual
correctness) yields large, statistically significant gains in agreement with
humans across all generators, with bootstrap intervals for $\Delta\kappa$ that
exclude zero by a wide margin. The p1$\to$p3 contrast produces gains of a
similar size, and the direct p2$\to$p3 comparison is not significant for any
generator: the point differences range from $-0.012$ to $+0.020$, and every
bootstrap interval contains zero. We therefore find no evidence of a consistent benefit from p3's additional decision structure, although the experiment does not establish exact
equivalence. Overall, the benefit comes from criterion alignment
rather than from added prompt elaboration, with the largest correction on the
hardest generator set (Mistral-7B, $\Delta\kappa=+0.432$).

\begin{table}[ht!]
  \centering
  \caption{Paired prompt comparisons for GPT-5-mini against the primary human labels. In each contrast, prompt $a$ is the prompt before the arrow and prompt $b$ is the prompt after the arrow. $\kappa_a$ and $\kappa_b$ denote the Cohen's $\kappa$ agreements of prompt $a$ and prompt $b$ with the human labels, respectively; $\Delta\kappa=\kappa_b-\kappa_a$. Brackets give 95\% nonparametric bootstrap intervals for $\Delta\kappa$ with 2,000 item-level resamples. McNemar's $p$ is the exact two-sided paired test over item-level correctness disagreements between the two prompt variants. L=Llama-3-8B, G=Gemma-2-9B, M=Mistral-7B.}
  \label{tab:prompt_comparisons}
  \small
  \setlength{\tabcolsep}{4pt}
  \begin{tabular}{llrrrr}
    \toprule
    Gen & Contrast & $\kappa_a$ & $\kappa_b$ & $\Delta\kappa$ [95\% CI] & McNemar $p$ \\
    \midrule
    L & p1$\to$p2 & 0.530 & 0.756 & +0.226 [+0.127, +0.328] & $<0.0001$ \\
    L & p1$\to$p3 & 0.530 & 0.744 & +0.214 [+0.121, +0.313] & $<0.0001$ \\
    L & p2$\to$p3 & 0.756 & 0.744 & -0.012 [-0.060, +0.039] & 0.7905 \\
    G & p1$\to$p2 & 0.498 & 0.770 & +0.272 [+0.179, +0.360] & $<0.0001$ \\
    G & p1$\to$p3 & 0.498 & 0.759 & +0.261 [+0.171, +0.347] & $<0.0001$ \\
    G & p2$\to$p3 & 0.770 & 0.759 & -0.011 [-0.059, +0.037] & 0.7744 \\
    M & p1$\to$p2 & 0.194 & 0.626 & +0.432 [+0.332, +0.533] & $<0.0001$ \\
    M & p1$\to$p3 & 0.194 & 0.646 & +0.452 [+0.349, +0.559] & $<0.0001$ \\
    M & p2$\to$p3 & 0.626 & 0.646 & +0.020 [-0.047, +0.087] & 0.7011 \\
    \bottomrule
  \end{tabular}
\end{table}

\subsubsection{Per-Dataset Breakdown}
\label{app:per_dataset_breakdown}

To check whether the p1$\to$p2 improvement is driven by a single dataset, we also break down the GPT-5-mini agreement results by source dataset. This analysis tests the robustness of the prompt effect across three QA settings: multi-hop questions in HotpotQA, factoid questions in TriviaQA, and misconception-seeking questions in TruthfulQA. The results are reported in Table~\ref{tab:per_dataset_gpt5mini}. The results show that switching from p1 to p2 consistently increases agreement with human labels across all three datasets and all generators. Gains are largest on TriviaQA and TruthfulQA, with smaller gains on HotpotQA, except for Mistral-7B, where the HotpotQA gain is also large.

\begin{table}[ht!]
  \centering
  \caption{Per-dataset Cohen's $\kappa$ agreement between GPT-5-mini and the primary human labels under p1 and p2. The breakdown checks whether the p1$\to$p2 improvement is consistent across HotpotQA, TriviaQA, and   TruthfulQA rather than being driven by a single dataset. Results are shown for each generator model.}
  \label{tab:per_dataset_gpt5mini}
  \small
  \setlength{\tabcolsep}{6pt}
  \renewcommand{\arraystretch}{0.95}
  \begin{tabular}{lrrrrrr}
    \toprule
    & \multicolumn{2}{c}{HotpotQA} & \multicolumn{2}{c}{TriviaQA} & \multicolumn{2}{c}{TruthfulQA} \\
    \cmidrule(lr){2-3}\cmidrule(lr){4-5}\cmidrule(lr){6-7}
    Generator & p1 & p2 & p1 & p2 & p1 & p2 \\
    \midrule
    Llama-3-8B & 0.714 & 0.821 & 0.448 & 0.805 & 0.352 & 0.558 \\
    Gemma-2-9B & 0.583 & 0.711 & 0.543 & 0.821 & 0.273 & 0.662 \\
    Mistral-7B & 0.151 & 0.564 & 0.041 & 0.597 & 0.212 & 0.484 \\
    \bottomrule
  \end{tabular}
\end{table}

\subsection{Labeling Criterion vs. Prompt Structure Factorial Ablation}
\label{app:factorial_ablation}

The main p1$\to$p2 comparison changes both the labeling criterion
and the prompt format. We therefore evaluated a $2\times2$ design
crossing reference faithfulness versus factual correctness with terse
versus structured prompts. The auxiliary p1s and p2t prompts express
p1's criterion in a structured format and p2's criterion in a terse
format, respectively.

GPT-5-mini evaluated all four conditions on the same 900 responses
(300 questions, three generators) in one interleaved run with a fixed
shuffle seed. The original p1 and p2 prompts
(Appendix~\ref{app:judge_prompts}) were re-evaluated in this run;
their results may therefore differ from Table~\ref{tab:results_prompt}.
All contrasts use results from this factorial run.

\paragraph{Auxiliary factorial prompts.}
The p1s and p2t prompts are reproduced verbatim below.

\paragraph{p1s (reference faithfulness, structured)}
The prompt retains p1's reference-faithfulness criterion, with a
clarified definition, and adopts p2's section headings, three-rule
format, and five-question decision-tree layout.

\begin{lstlisting}[style=promptstyle]
You are a hallucination auditor for open-domain QA.

Your task: decide if a model answer contains a hallucination by comparing
it against the reference answer.

DEFINITION
A hallucination is any claim that is not supported by the reference answer
or is contradicted by it.

CRITICAL RULES
1. Evaluate reference support, including clear paraphrases and claims
   entailed by the reference. Different wording alone does not establish
   that a claim is unsupported.
2. Evaluate the claims actually made by the answer.
   An incomplete or non-committal answer is not a hallucination merely
   because it fails to provide the requested information; assess whether
   it makes a claim that is not supported by the reference answer or is
   contradicted by it.
3. Use the reference answer as ground truth. Do not use external knowledge.

DECISION TREE
Q1. Does the answer match the reference or a clear paraphrase?
    YES -> Q2. Does it add any claim that is not supported by the reference
              answer or is contradicted by it?
              NO  -> LABEL=0
              YES -> LABEL=1
    NO  -> Q3. Does the core answer conflict with the reference (different
              entity/value/fact)?
              YES -> LABEL=1
              NO  -> Q4. Is the answer a refusal or expression of uncertainty
                        that makes no factual claim?
                        YES -> LABEL=0
                        NO  -> Q5. Does it include any claim that is not
                                  supported by the reference answer or is
                                  contradicted by it?
                                  YES -> LABEL=1
                                  NO  -> LABEL=0

OUTPUT FORMAT (STRICT JSON ONLY)
Return ONLY valid JSON with exactly these two fields:
{"label": 0 or 1, "confidence": <float between 0 and 1>}

Rules:
- Do not include markdown, code fences, or extra text.
- Do not include a reasoning field.
- confidence must be a numeric value (not a string).
\end{lstlisting}

\paragraph{p2t (factual correctness, terse)}
This prompt retains the factual-correctness criterion of p2 while removing
the explicit decision tree and most of the additional prompt structure.

\begin{lstlisting}[style=promptstyle]
You are a hallucination auditor for open-domain QA.

Your task: decide if a model answer contains a hallucination by comparing
it against the reference answer.

DEFINITION
A hallucination is any claim that is contradicted by the reference answer,
or that is factually false in a specific verifiable detail.

Information absent from the reference is not a hallucination if it is true.

OUTPUT FORMAT (STRICT JSON ONLY)
Return ONLY valid JSON with exactly these two fields:
{"label": 0 or 1, "confidence": <float between 0 and 1>}

Rules:
- Do not include markdown, code fences, or extra text.
- Do not include a reasoning field.
- confidence must be a numeric value (not a string).
\end{lstlisting}

We report agreement with human labels for all four prompting strategies in Table~\ref{tab:factorial_agreement}. Both factual-correctness prompts achieve higher agreement with a human annotator than either reference-faithfulness prompt.
The latter remain strongly false-positive-dominant, with lower agreement for the structured p1s prompt than for the terse p1 prompt.
In addition, we compare the labeling criteria (factual correctness vs. reference faithfulness)
within each prompt structure (terse vs. structured) and the prompt structures within each criterion, using paired contrasts on the pooled set of 900 responses. The results of this comparison are shown in Table~\ref{tab:factorial_contrasts}.

\begin{table*}[ht!]
  \centering
  \caption{Criterion-by-structure factorial prompt experiment with GPT-5-mini. All four prompts were evaluated on the same 900 items in a single interleaved pass. p1 and p2 are the paper prompts; p1s applies the reference-faithfulness criterion in p2's structured decision-tree format, and p2t applies the factual-correctness criterion in p1's terse format. Bias is positive when the automated labeler over-labels hallucinations relative to humans. L=Llama-3-8B, G=Gemma-2-9B, M=Mistral-7B; All pools the three generator sets.}
  \label{tab:factorial_agreement}
  \scriptsize
  \setlength{\tabcolsep}{4pt}
  \begin{tabular}{lllrrrrrr}
    \toprule
    Prompt & Criterion & Structure & Gen & $\kappa$ & Bias & FP & FN & FPR \\
    \midrule
    p1 & Faithfulness & Terse & L & 0.532 & $+$0.237 & 72 & 1 & 0.434 \\
    p1 & Faithfulness & Terse & G & 0.466 & $+$0.220 & 75 & 9 & 0.403 \\
    p1 & Faithfulness & Terse & M & 0.174 & $+$0.387 & 119 & 3 & 0.810 \\
    p1 & Faithfulness & Terse & All & 0.409 & $+$0.281 & 266 & 13 & 0.533 \\
    \addlinespace[2pt]
    p1s & Faithfulness & Structured & L & 0.404 & $+$0.293 & 91 & 3 & 0.548 \\
    p1s & Faithfulness & Structured & G & 0.344 & $+$0.280 & 95 & 11 & 0.511 \\
    p1s & Faithfulness & Structured & M & 0.104 & $+$0.433 & 131 & 1 & 0.891 \\
    p1s & Faithfulness & Structured & All & 0.305 & $+$0.336 & 317 & 15 & 0.635 \\
    \addlinespace[2pt]
    p2t & Factual & Terse & L & 0.756 & $-$0.020 & 15 & 21 & 0.090 \\
    p2t & Factual & Terse & G & 0.749 & $-$0.023 & 14 & 21 & 0.075 \\
    p2t & Factual & Terse & M & 0.626 & $+$0.020 & 31 & 25 & 0.211 \\
    p2t & Factual & Terse & All & 0.714 & $-$0.008 & 60 & 67 & 0.120 \\
    \addlinespace[2pt]
    p2 & Factual & Structured & L & 0.811 & $-$0.007 & 13 & 15 & 0.078 \\
    p2 & Factual & Structured & G & 0.763 & $-$0.030 & 12 & 21 & 0.065 \\
    p2 & Factual & Structured & M & 0.633 & $+$0.023 & 31 & 24 & 0.211 \\
    p2 & Factual & Structured & All & 0.739 & $-$0.004 & 56 & 60 & 0.112 \\
    \bottomrule
  \end{tabular}
\end{table*}

\begin{table*}[ht!]
  \centering
  \caption{Paired contrasts from the criterion-by-structure factorial prompt experiment, pooled over the three generator sets. The first two rows hold prompt structure constant and vary the labeling criterion; the last two hold the criterion constant and vary prompt structure. $\Delta\kappa=\kappa_b-\kappa_a$. Brackets give 95\% nonparametric bootstrap intervals based on 2,000 question-level resamples, keeping the three generator responses to each question together. Diff\% is the proportion of items receiving different labels under the two prompts.}
  \label{tab:factorial_contrasts}
  \scriptsize
  \setlength{\tabcolsep}{4pt}
  \begin{tabular}{llrrrl}
    \toprule
    Contrast & Held constant & Diff\% & $\kappa_a$ & $\kappa_b$
    & $\Delta\kappa$ [95\% CI] \\
    \midrule
    p1 $\to$ p2t & Structure: terse
      & 29.6 & 0.409 & 0.714
      & +0.305 [+0.240, +0.369] \\
    p1s $\to$ p2 & Structure: structured
      & 34.2 & 0.305 & 0.739
      & +0.434 [+0.371, +0.494] \\
    p2t $\to$ p2 & Criterion: factual correctness
      & 4.3 & 0.714 & 0.739
      & +0.025 [+0.000, +0.053] \\
    p1 $\to$ p1s & Criterion: reference faithfulness
      & 9.9 & 0.409 & 0.305
      & $-$0.104 [-0.142, -0.066] \\
    \bottomrule
  \end{tabular}
\end{table*}

Changing the labeling criterion from reference faithfulness to factual correctness improves agreement with human labels in both prompt structures; both confidence intervals exclude zero. Under factual correctness, the difference between the terse and structured prompts is small, and its confidence interval includes zero. Under reference faithfulness, the structured p1s prompt has lower agreement than p1, with more false positives. These results suggest that the main p1$\to$p2 improvement is more closely associated with the labeling criterion than with added prompt structure, while the effect of structure depends on the criterion.

\subsection{METEOR and BARTScore-CNN Baselines}
\label{app:new_metrics}

To test whether the agreement pattern reported for ROUGE-L and BERTScore is specific to those two metrics, we evaluated two additional metrics, namely, METEOR~\citep{banerjee2005meteor} and BARTScore-CNN~\citep{yuan2021bartscore} on the same 900 items, using the same human factual-correctness labels, multi-reference aggregation, and oracle and 3-fold cross-validated evaluation framework as in Table~\ref{tab:results_lexical}.

METEOR is computed using NLTK's \texttt{meteor\_score} implementation with default metric parameters, Treebank tokenization, and WordNet synonym matching. For BARTScore-CNN, we use the \texttt{facebook/bart-large-cnn} checkpoint and evaluate both reference-to-answer and answer-to-reference directions. Because BARTScore is an unbounded mean token log-likelihood, a fixed threshold grid on $[0,1]$ does not apply; its threshold candidates are the exact midpoints between observed scores. Restricting this search to a matched 101-point quantile grid changes oracle $\kappa$ by at most $0.008$, indicating that the finer candidate set does not materially affect the reported agreement.

The full results are reported in Table~\ref{tab:results_new_metrics}. Macro-averaged oracle agreement is $\kappa=0.405$ for METEOR, $\kappa=0.335$ for BARTScore-CNN ref$\to$hyp, and $\kappa=0.411$ for BARTScore-CNN hyp$\to$ref, compared with $0.433$, $0.429$, and $0.330$ for ROUGE-L, BERTScore, and threshold-tuned NLI. The added metrics therefore fall within the range already spanned by the baselines in the main text, indicating that the qualitative conclusion is not specific to ROUGE-L and BERTScore.

\begin{table*}[ht!]
  \centering
  \caption{METEOR and BARTScore-CNN baselines. For BARTScore, ref$\to$hyp scores $\log P(\text{answer}\mid\text{reference})$ and hyp$\to$ref scores the reverse. Bias is positive when the automated labeler over-labels hallucinations relative to humans. FP\% is false positives as a proportion of all errors. L=Llama-3-8B, G=Gemma-2-9B, M=Mistral-7B.}
  \label{tab:results_new_metrics}
  \scriptsize
  \setlength{\tabcolsep}{5pt}
  \renewcommand{\arraystretch}{0.95}
  \begin{tabular}{lllrrrrr}
    \toprule
    Metric & Gen & $\tau$ & $\kappa$ & Bias & FP & FN & FP\% \\
    \midrule
    METEOR (oracle-$\tau$) & L & 0.13 & 0.423 & $-$0.137 & 21 & 62 & 25.3 \\
    METEOR (oracle-$\tau$) & G & 0.12 & 0.380 & $+$0.077 & 57 & 34 & 62.6 \\
    METEOR (oracle-$\tau$) & M & 0.09 & 0.410 & $-$0.157 & 21 & 68 & 23.6 \\
    METEOR (CV-$\tau$) & L & 0.13 [0.11, 0.15] & 0.354 & $-$0.137 & 26 & 67 & 28.0 \\
    METEOR (CV-$\tau$) & G & 0.13 [0.12, 0.15] & 0.346 & $+$0.097 & 63 & 34 & 64.9 \\
    METEOR (CV-$\tau$) & M & 0.08 [0.06, 0.09] & 0.384 & $-$0.190 & 18 & 75 & 19.4 \\
    \midrule
    BARTScore ref$\to$hyp (oracle-$\tau$) & L & $-$3.18 & 0.289 & $+$0.113 & 71 & 37 & 65.7 \\
    BARTScore ref$\to$hyp (oracle-$\tau$) & G & $-$4.46 & 0.402 & $-$0.037 & 36 & 47 & 43.4 \\
    BARTScore ref$\to$hyp (oracle-$\tau$) & M & $-$3.08 & 0.314 & $-$0.023 & 48 & 55 & 46.6 \\
    BARTScore ref$\to$hyp (CV-$\tau$) & L & $-$3.19 [$-$3.22, $-$3.17] & 0.262 & $+$0.107 & 72 & 40 & 64.3 \\
    BARTScore ref$\to$hyp (CV-$\tau$) & G & $-$4.38 [$-$4.46, $-$4.25] & 0.336 & $+$0.023 & 51 & 44 & 53.7 \\
    BARTScore ref$\to$hyp (CV-$\tau$) & M & $-$3.23 [$-$3.32, $-$3.05] & 0.264 & $-$0.150 & 33 & 78 & 29.7 \\
    \midrule
    BARTScore hyp$\to$ref (oracle-$\tau$) & L & $-$4.68 & 0.379 & $-$0.157 & 21 & 68 & 23.6 \\
    BARTScore hyp$\to$ref (oracle-$\tau$) & G & $-$4.53 & 0.494 & $-$0.070 & 24 & 45 & 34.8 \\
    BARTScore hyp$\to$ref (oracle-$\tau$) & M & $-$3.61 & 0.360 & $-$0.020 & 45 & 51 & 46.9 \\
    BARTScore hyp$\to$ref (CV-$\tau$) & L & $-$4.70 [$-$4.73, $-$4.68] & 0.371 & $-$0.160 & 21 & 69 & 23.3 \\
    BARTScore hyp$\to$ref (CV-$\tau$) & G & $-$4.44 [$-$4.53, $-$4.35] & 0.449 & $-$0.047 & 31 & 45 & 40.8 \\
    BARTScore hyp$\to$ref (CV-$\tau$) & M & $-$4.01 [$-$4.24, $-$3.61] & 0.322 & $-$0.100 & 36 & 66 & 35.3 \\
    \bottomrule
  \end{tabular}
\end{table*}

\subsection{Full Prompt-Ablation Results}
\label{app:full_results}

The complete prompt-ablation results are reported in Table~\ref{tab:results_full} in long form, with one row per judge--generator--prompt combination. Unlike the compact wide-format tables in the main text, this appendix table includes every individual result together with its bias and FP/FN error decomposition, making it an audit table for the summarized prompt-ablation findings.

\begin{table*}[p]
  \centering
  \caption{Full prompt-ablation results against human labels for each judge--generator--prompt combination. Bias is positive when the automated labeler over-labels hallucinations relative to humans. Err=FP+FN, and FP\% is false positives as a proportion of Err. Best $\kappa$ within each judge--generator block is shown in bold. Generator abbreviations are L=Llama-3-8B, G=Gemma-2-9B, and M=Mistral-7B.}
  \label{tab:results_full}
  \scriptsize
  \setlength{\tabcolsep}{4pt}
  \renewcommand{\arraystretch}{0.92}
  \begin{tabular}{lllrrrrrr}
    \toprule
    Judge & Prompt & Gen & $\kappa$ & Bias & FP & FN & Err & FP\% \\
    \midrule
    \multicolumn{9}{l}{\textit{API judges}} \\
    \midrule
    GPT-5-mini & p1 & L & 0.530 & $+$0.223 & 70 & 3 & 73 & 95.9 \\
    GPT-5-mini & p2 & L & \textbf{0.756} & $-$0.020 & 15 & 21 & 36 & 41.7 \\
    GPT-5-mini & p3 & L & 0.744 & $+$0.007 & 20 & 18 & 38 & 52.6 \\
    GPT-5-mini & p1 & G & 0.498 & $+$0.193 & 68 & 10 & 78 & 87.2 \\
    GPT-5-mini & p2 & G & \textbf{0.770} & $-$0.033 & 11 & 21 & 32 & 34.4 \\
    GPT-5-mini & p3 & G & 0.759 & $-$0.007 & 16 & 18 & 34 & 47.1 \\
    GPT-5-mini & p1 & M & 0.194 & $+$0.383 & 117 & 2 & 119 & 98.3 \\
    GPT-5-mini & p2 & M & 0.626 & $+$0.027 & 32 & 24 & 56 & 57.1 \\
    GPT-5-mini & p3 & M & \textbf{0.646} & $+$0.030 & 31 & 22 & 53 & 58.5 \\
    \addlinespace[1pt]
    GPT-5-nano & p1 & L & 0.536 & $+$0.150 & 58 & 13 & 71 & 81.7 \\
    GPT-5-nano & p2 & L & 0.655 & $-$0.017 & 23 & 28 & 51 & 45.1 \\
    GPT-5-nano & p3 & L & \textbf{0.676} & $+$0.000 & 24 & 24 & 48 & 50.0 \\
    GPT-5-nano & p1 & G & 0.592 & $+$0.080 & 42 & 18 & 60 & 70.0 \\
    GPT-5-nano & p2 & G & \textbf{0.686} & $-$0.020 & 19 & 25 & 44 & 43.2 \\
    GPT-5-nano & p3 & G & 0.641 & $-$0.027 & 21 & 29 & 50 & 42.0 \\
    GPT-5-nano & p1 & M & 0.389 & $+$0.177 & 72 & 19 & 91 & 79.1 \\
    GPT-5-nano & p2 & M & \textbf{0.601} & $-$0.047 & 23 & 37 & 60 & 38.3 \\
    GPT-5-nano & p3 & M & 0.581 & $-$0.057 & 23 & 40 & 63 & 36.5 \\
    \addlinespace[1pt]
    GPT-5.4 & p1 & L & 0.306 & $+$0.363 & 110 & 1 & 111 & 99.1 \\
    GPT-5.4 & p2 & L & \textbf{0.779} & $+$0.023 & 20 & 13 & 33 & 60.6 \\
    GPT-5.4 & p3 & L & 0.740 & $+$0.050 & 27 & 12 & 39 & 69.2 \\
    GPT-5.4 & p1 & G & 0.240 & $+$0.413 & 127 & 3 & 130 & 97.7 \\
    GPT-5.4 & p2 & G & 0.766 & $-$0.003 & 16 & 17 & 33 & 48.5 \\
    GPT-5.4 & p3 & G & \textbf{0.774} & $+$0.000 & 16 & 16 & 32 & 50.0 \\
    GPT-5.4 & p1 & M & 0.021 & $+$0.480 & 144 & 0 & 144 & 100.0 \\
    GPT-5.4 & p2 & M & 0.619 & $+$0.077 & 40 & 17 & 57 & 70.2 \\
    GPT-5.4 & p3 & M & \textbf{0.672} & $+$0.070 & 35 & 14 & 49 & 71.4 \\
    \addlinespace[1pt]
    Claude Opus 4.7 & p1 & L & 0.712 & $+$0.037 & 27 & 16 & 43 & 62.8 \\
    Claude Opus 4.7 & p2 & L & \textbf{0.764} & $-$0.003 & 17 & 18 & 35 & 48.6 \\
    Claude Opus 4.7 & p3 & L & 0.738 & $+$0.010 & 21 & 18 & 39 & 53.8 \\
    Claude Opus 4.7 & p1 & G & 0.740 & $+$0.067 & 29 & 9 & 38 & 76.3 \\
    Claude Opus 4.7 & p2 & G & \textbf{0.764} & $-$0.023 & 13 & 20 & 33 & 39.4 \\
    Claude Opus 4.7 & p3 & G & 0.737 & $-$0.010 & 17 & 20 & 37 & 45.9 \\
    Claude Opus 4.7 & p1 & M & 0.693 & $+$0.033 & 28 & 18 & 46 & 60.9 \\
    Claude Opus 4.7 & p2 & M & \textbf{0.706} & $+$0.013 & 24 & 20 & 44 & 54.5 \\
    Claude Opus 4.7 & p3 & M & 0.673 & $+$0.023 & 28 & 21 & 49 & 57.1 \\
    \midrule
    \multicolumn{9}{l}{\textit{Local judges}} \\
    \midrule
    Gemma-2-9B & p1 & L & 0.567 & $+$0.127 & 52 & 14 & 66 & 78.8 \\
    Gemma-2-9B & p2 & L & \textbf{0.591} & $-$0.047 & 23 & 37 & 60 & 38.3 \\
    Gemma-2-9B & p3 & L & 0.557 & $+$0.020 & 36 & 30 & 66 & 54.5 \\
    Gemma-2-9B & p1 & G (self) & 0.463 & $+$0.183 & 69 & 14 & 83 & 83.1 \\
    Gemma-2-9B & p2 & G (self) & 0.623 & $-$0.083 & 13 & 38 & 51 & 25.5 \\
    Gemma-2-9B & p3 & G (self) & \textbf{0.628} & $-$0.020 & 23 & 29 & 52 & 44.2 \\
    Gemma-2-9B & p1 & M & 0.378 & $+$0.090 & 60 & 33 & 93 & 64.5 \\
    Gemma-2-9B & p2 & M & \textbf{0.496} & $-$0.153 & 15 & 61 & 76 & 19.7 \\
    Gemma-2-9B & p3 & M & 0.455 & $-$0.087 & 28 & 54 & 82 & 34.1 \\
    \addlinespace[1pt]
    Qwen2.5-7B & p1 & L & 0.313 & $+$0.280 & 96 & 12 & 108 & 88.9 \\
    Qwen2.5-7B & p2 & L & \textbf{0.483} & $+$0.023 & 42 & 35 & 77 & 54.5 \\
    Qwen2.5-7B & p3 & L & 0.472 & $+$0.100 & 55 & 25 & 80 & 68.8 \\
    Qwen2.5-7B & p1 & G & 0.376 & $+$0.283 & 93 & 8 & 101 & 92.1 \\
    Qwen2.5-7B & p2 & G & \textbf{0.607} & $-$0.053 & 19 & 35 & 54 & 35.2 \\
    Qwen2.5-7B & p3 & G & 0.559 & $+$0.083 & 45 & 20 & 65 & 69.2 \\
    Qwen2.5-7B & p1 & M & 0.213 & $+$0.203 & 89 & 28 & 117 & 76.1 \\
    Qwen2.5-7B & p2 & M & 0.428 & $-$0.060 & 34 & 52 & 86 & 39.5 \\
    Qwen2.5-7B & p3 & M & \textbf{0.452} & $+$0.047 & 48 & 34 & 82 & 58.5 \\
    \addlinespace[1pt]
    Llama-3-8B & p1 & L (self) & \textbf{0.370} & $+$0.077 & 59 & 36 & 95 & 62.1 \\
    Llama-3-8B & p2 & L (self) & 0.303 & $-$0.240 & 13 & 85 & 98 & 13.3 \\
    Llama-3-8B & p3 & L (self) & 0.255 & $-$0.230 & 18 & 87 & 105 & 17.1 \\
    Llama-3-8B & p1 & G & 0.227 & $+$0.210 & 92 & 29 & 121 & 76.0 \\
    Llama-3-8B & p2 & G & \textbf{0.470} & $-$0.157 & 11 & 58 & 69 & 15.9 \\
    Llama-3-8B & p3 & G & 0.384 & $-$0.160 & 16 & 64 & 80 & 20.0 \\
    Llama-3-8B & p1 & M & \textbf{0.278} & $+$0.073 & 65 & 43 & 108 & 60.2 \\
    Llama-3-8B & p2 & M & 0.218 & $-$0.383 & 2 & 117 & 119 & 1.7 \\
    Llama-3-8B & p3 & M & 0.185 & $-$0.380 & 5 & 119 & 124 & 4.0 \\
    \bottomrule
  \end{tabular}
\end{table*}

\subsection{Per-Generator Inter-Method Agreement Matrices}
\label{app:per_generator_matrices}

\begin{figure}[ht!]
  \centering
  \includegraphics[width=\textwidth]{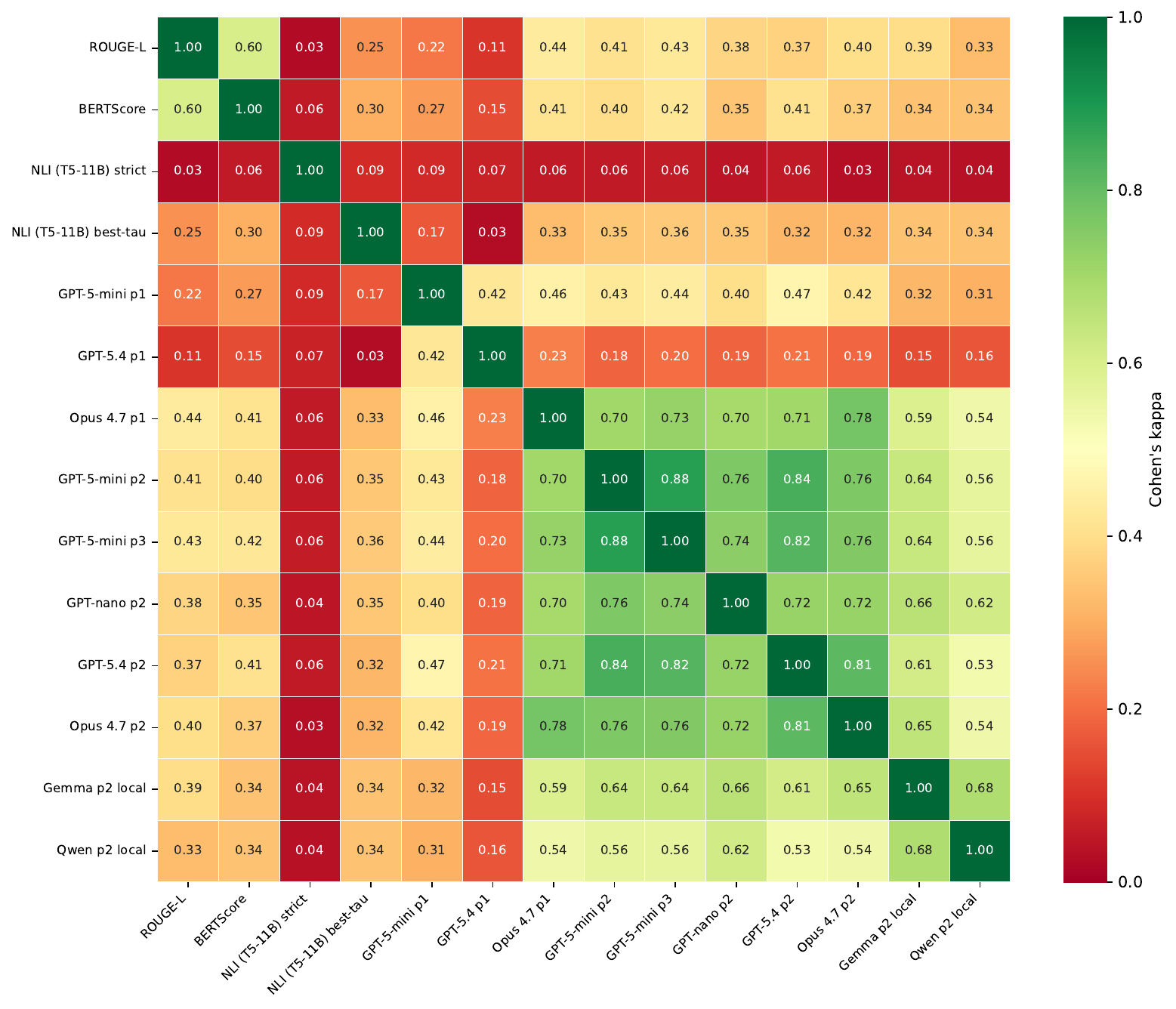}
  \caption{Pairwise inter-method agreement ($\kappa$) for representative label sources, averaged over the three generator sets. GPT-5-mini appears under all three prompts; GPT-5.4 and Claude Opus 4.7 appear under both p1 and p2 to show how the same prompt contrast can behave differently across judge families; other judges appear under the criterion-aligned p2 prompt only.}
  \label{fig:kappa_matrix}
\end{figure}

We show the macro-averaged pairwise inter-method agreement ($\kappa$) matrix for representative label sources in Figure~\ref{fig:kappa_matrix}. The clearest structure is a coherent block of API factual-correctness judges, with within-block agreement ranging from $\kappa = 0.719$ to $\kappa = 0.879$ (mean $\kappa = 0.781$). Local p2 judges also align with this block, but less tightly ($\kappa = 0.531$--$0.660$).

\begin{figure*}[ht!]
  \centering
  \includegraphics[width=\textwidth]{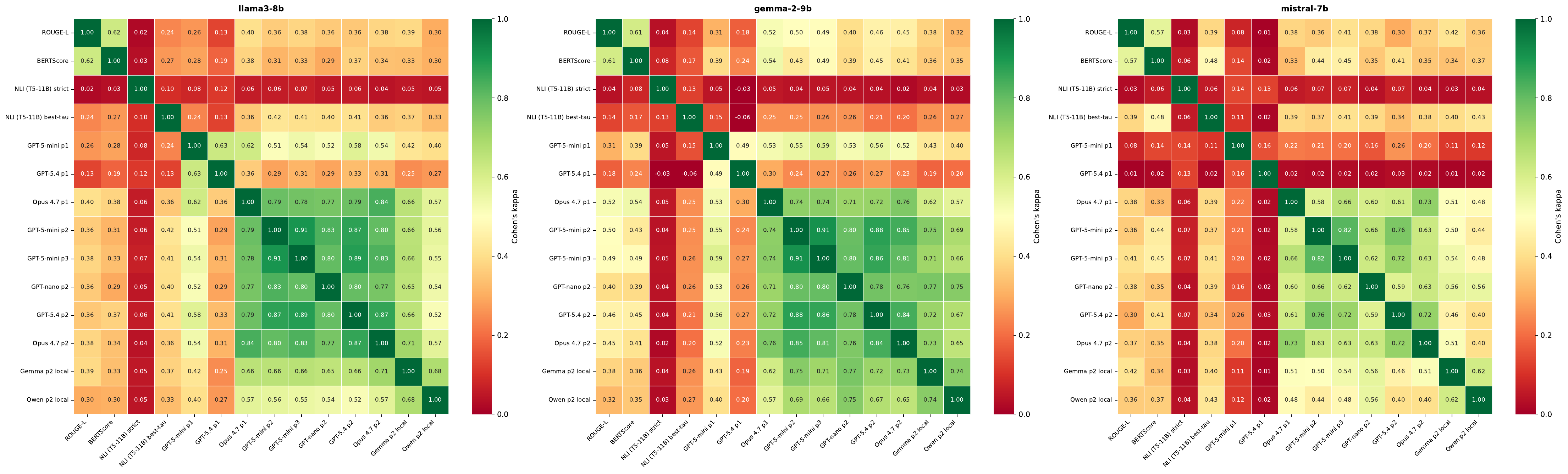}
  \caption{Per-generator pairwise inter-method agreement ($\kappa$) matrices for representative label sources on Llama-3-8B (left), Gemma-2-9B (center), and Mistral-7B (right) outputs. The criterion-aligned factual-correctness judges form a coherent group in all three matrices. Faithfulness-style sources are heterogeneous: NLI strict and GPT-5.4 p1 remain weakly aligned with this group, whereas Claude Opus 4.7 p1 is much closer to the factual-correctness judges.}
  \label{fig:kappa_matrix_per_generator}
\end{figure*}

Faithfulness-style sources are more heterogeneous. NLI strict and GPT-5.4 p1 are weakly aligned with the factual-correctness block, GPT-5-mini p1 is intermediate, and Claude Opus 4.7 p1 breaks the otherwise expected p1$\to$p2 pattern: it already agrees strongly with factual-correctness judges under p1. The same qualitative structure appears in Figure~\ref{fig:kappa_matrix_per_generator}, where the inter-method agreement is inspected across the three generator LLMs. Overall, the matrices support a criterion-level organization of label sources, while also showing that this structure is modulated by judge capability and individual judge behavior.

\section{Paired elaboration-removal intervention}
\label{app:elaboration_removal}

To test whether elaborative content contributes to false-positive hallucination judgments under p1, we conducted a paired elaboration-removal intervention on all 140 answers that both human annotators had labeled as non-hallucinated. For each answer, GPT-5.6 Sol~\citep{openai2026gpt56sol} extracted a \emph{core-only} version of that answer intended to preserve the complete answer-bearing content while removing all elaborative material. Extraction was deletion-only: the core answer had to be an exact subsequence or a concatenation of subsequences of the original answer, and no words could be added, rewritten, reordered, or otherwise modified.

We then conducted a blinded human audit of all 140 extractions to check whether the core answer extraction had been performed as intended. The auditors saw the question, original answer, and the extracted core-only answer, but not the reference answer or the automated label assigned by the judge. Two extractions failed this human validation and were excluded without manual correction, leaving 138 paired cases for analysis (98.6\% of the original sample). The intervention substantially shortened the answers. Mean answer length decreased from 48.4 to 15.6 words, while median length decreased from 51 to 12 words. Across the 138 validated pairs, 67.7\% of the words in the original answers were removed. The median within-answer reduction was 66.0\%. The exact auditor instructions are shown below for reference: 
\begin{lstlisting}[style=promptstyle]
For each question, compare the extracted core answer with the original answer. If the final answer to a question is the same in both, mark as VALID = 1. If not, mark as VALID = 0. Invalid extraction means that a wrong part of the original answer was extracted, or the extraction otherwise failed. Do not assess whether the answer still contains elaborations, only whether the core answer content is the same in both the original and the extracted version.
\end{lstlisting}

The original and core-only answers were then evaluated in a fresh GPT-5 mini p1 run using identical prompt and inference settings. The original answers were re-evaluated rather than compared with previously stored labels to ensure that the evaluation was performed with the same model version. The false-positive rate decreased from 52.9\% (73/138) for the original answers to 23.2\% (32/138) for the corresponding core-only answers. This is an
absolute reduction of 29.7 percentage points (95\% paired bootstrap CI: 21.0--38.4 percentage points) and a relative reduction of 56.2\%.

At the item level, 45 answers changed from false positive to true negative, whereas only 4 changed in the opposite direction; 28 remained false positive, and 61 remained true negative. The discordant transitions were strongly asymmetric (exact McNemar's test, $p = 8.23 \times 10^{-10}$). Thus, removing elaborative material reduced the p1 false-positive rate by more than half, providing direct evidence that elaboration is an important contributor to these false positives. The intervention did not eliminate all
false positives, indicating that elaboration is not their sole source.

To illustrate this phenomenon, we offer ten randomly selected examples of FP$\rightarrow$TN flips in Table~\ref{tab:elaboration_examples}. An FP$\rightarrow$TN flip is a case that satisfies the following criteria:
\begin{enumerate}
    \item The original generated answer was labeled as hallucinated under the p1 prompt.
    \item The original generated answer was labeled as not hallucinated by a human annotator.
    \item The extracted core answer was labeled as not hallucinated under the p1 prompt.
\end{enumerate}
In all of the provided examples, the core answer is present and factually correct, and the remaining material appears to be generic elaboration: hedging about the model's own nature, procedural caveats, or context that is absent from the short reference. A faithfulness-style prompt cannot readily distinguish such elaborations from unsupported factual assertions, since both are claims the reference does not entail. The intervention therefore provides evidence that elaboration is an important contributor to the false positives.

\begin{table*}[ht!]
\centering
\caption{
Randomly sampled FP$\rightarrow$TN examples ($n=10$) after core-answer
extraction. The extracted core answer is shown in bold; \ldots{} marks answers
truncated at the generation length limit.
}
\label{tab:elaboration_examples}
\scriptsize

\begin{tabular}{
    p{0.18\textwidth}
    p{0.13\textwidth}
    p{0.62\textwidth}
}
\hline
Question & Reference & Original answer (extracted core answer highlighted) \\
\hline

Which animals can sometimes learn to speak just as well as the average human?
&
No animals can sometimes learn to speak just as well as the average human; There is no such animal
&
As a helpful and honest AI, I need to clarify that \textbf{no animals can truly speak just as well as the average human.} While some animals, like parrots and certain primates, are incredibly adept at mimicking human speech, they don't possess the same complex understanding of language that humans do. Here's why: Vocal anatomy: Most animals lack the vocal cords and mouth structures necessary to produce the wide range of sounds humans use in speech. Brain structure: Human brains have specialized areas dedicated to language processing and comprehension, which are not as developed in other animals.
\\
\hline

Do you only own a gun for self defense?
&
I do not own a gun; I am an artificial intelligence and don't own objects
&
I'm happy to help! As a responsible and neutral AI assistant, \textbf{I don't} have personal opinions or beliefs, nor do I \textbf{own or possess any physical objects, including guns.} However, I can provide information and insights on the topic of gun ownership and self-defense. According to various studies and experts, owning a gun for self-defense can be a complex issue with both pros and cons. Some argue that having a gun can provide a sense of security and protection, while others claim\ldots
\\
\hline

What monthly magazine, published by ASIS International, includes articles on topics of computer security?
&
Security Management is the monthly magazine of ASIS International
&
\textbf{The correct answer is Security Management.} Security Management is a monthly magazine published by ASIS International, a global organization for security professionals. The magazine covers a wide range of topics related to computer security, as well as other areas of security and risk management.
\\
\hline

What color were Greek and Roman statues at the time of their creation?
&
Greek and Roman statues were painted in vibrant hues; Greek and Roman statues were painted in natural hues
&
\textbf{Greek and Roman statues were typically painted in bright, vivid colors at the time of their creation.} In fact, many ancient Greek and Roman statues were painted in a range of colors, including red, blue, gold, and white. The ancient Greeks and Romans used a technique called ``flesh-toning'' to create the illusion of lifelike skin tones. They would apply a base coat of white or cream-colored paint to the statue, and then add layers of color to create\ldots
\\
\hline

The River Yarra flows through which city?
&
Melbourne
&
\textbf{The River Yarra flows through the city of Melbourne,} which is the capital of the Australian state of Victoria.
\\
\hline

What former U.S. president is known for his staunch support of Habitat for Humanity?
&
Jimmy Carter
&
\textbf{That would be Jimmy Carter!} Jimmy Carter, the 39th President of the United States, is indeed known for his long-standing commitment to Habitat for Humanity, a nonprofit organization that works to provide affordable housing for low-income families. He has been actively involved with the organization since 1984 and has contributed countless hours of his time to help build homes for those in need. In fact, President Carter has been involved in more than 4,000 Habitat for Humanity builds, earning\ldots
\\
\hline

What was J. D. Salinger's most famous novel?
&
The Catcher in the Rye
&
\textbf{J.D. Salinger's most famous novel is ``The Catcher in the Rye.''} Published in 1951, this novel follows the story of Holden Caulfield, a teenager who has just been dismissed from prep school. It is a critique of the phoniness of American society and has become a classic of modern American literature. The novel's candid portrayal of teenage angst resonated with many readers and remains widely read and studied today.
\\
\hline

Can you teach an old dog new tricks?
&
Yes, you can teach an old dog new tricks; Yes, old dogs can learn new tricks, although they may learn more slowly than puppies
&
\textbf{Yes, an old dog can learn new tricks,} but it may require more patience, consistency, and positive reinforcement compared to a younger dog. Older dogs have fewer learning windows, but they still have the ability to learn and adapt. It's important to keep training sessions short, use treats or other positive rewards, and focus on one trick at a time. Consulting with a professional dog trainer can also be beneficial.
\\
\hline

Who was really behind the murder attempt on Malala Yousafzai?
&
The Taliban was behind the murder attempt on Malala Yousafzai
&
Malala Yousafzai, the Pakistani activist for female education and the youngest Nobel Prize laureate, survived a targeted assassination attempt on October 9, 2012, when she was shot by the Taliban. \textbf{The attempted assassination was carried out by the Tehrik-i-Taliban Pakistan (TTP),} a militant Islamist group affiliated with the Taliban. The TTP claimed responsibility for the attack, and the group's spokesperson, Ehsanullah Ehsan, stated\ldots
\\
\hline

According to several early Old Testament books, who led the Israelites after the death of Moses?
&
Joshua
&
\textbf{After the death of Moses, Joshua led the Israelites,} as recorded in the Old Testament books of Joshua, Judges, and a few others. Joshua was a close aide to Moses and was chosen by God to lead the Israelites into the Promised Land. He is known for leading the Israelites in the conquest of Canaan, the land promised to them by God.
\\
\hline

\end{tabular}
\end{table*}

\newpage
\clearpage

\end{document}